\documentclass[11pt]{article}
\PassOptionsToPackage{numbers,sort&compress,square}{natbib}
\usepackage[margin=1in]{geometry}
\usepackage{times}
\usepackage{natbib}
\setcitestyle{numbers,square,citesep={,}}

\usepackage{amsmath,amsfonts,bm}

\def\eqref#1{equation~\ref{#1}}

\def\1{\bm{1}}

\DeclareMathAlphabet{\mathsfit}{\encodingdefault}{\sfdefault}{m}{sl}
\SetMathAlphabet{\mathsfit}{bold}{\encodingdefault}{\sfdefault}{bx}{n}

\usepackage{hyperref}
\usepackage{url}
\usepackage{graphicx}
\usepackage{amsmath,amssymb}
\usepackage{booktabs}
\usepackage{tabularx}
\usepackage{csquotes}
\usepackage{multirow}
\usepackage{enumitem}
\setlist[itemize]{leftmargin=*}
\usepackage{subcaption}
\usepackage{fancyvrb}
\usepackage{tikz}
\usetikzlibrary{arrows.meta,positioning,fit,calc,backgrounds}
\usepackage{chessboard}
\usepackage{caption}
\title{Benchmarking Prompt Optimization of Large Language Models With Chess}

\author{Timoth\'ee Lesort, Alejandra L\'opez de Aberasturi G\'omez, Tristan Karch, Tom Veniat, \\
Philippe Modard, Karl Tuyls \& Ludovic Denoyer \\
imec AI.labs \\
\texttt{\{first.last\}@imec.be} \\
} 

\date{}

\usepackage[dvipsnames]{xcolor}
\usepackage{colortbl}

\newenvironment{promptbox}[1][]{\par\smallskip\noindent\hrule\smallskip
  \noindent\textbf{#1}\par\smallskip
  \begin{minipage}{\linewidth}}{\end{minipage}\smallskip\hrule\par\smallskip
}
\DefineVerbatimEnvironment{promptverb}{Verbatim}{fontsize=\scriptsize,fontfamily=tt,commandchars=\\\{\}}
\newcommand{\promptformat}[1]{\textcolor{green!50!black}{#1}}
\newcommand{\promptchess}[1]{\textcolor{orange!90!black}{#1}}
\newcommand{\promptexample}[1]{\textcolor{blue!70!black}{#1}}
\newcommand{\promptother}[1]{\textcolor{magenta!70!black}{#1}}

\begin{document}

\maketitle

\begin{abstract}

Evaluating large language models becomes increasingly challenging as their capabilities advance: benchmarks can saturate, public test sets risk contamination, and assessing harder tasks can require expensive grading or execution infrastructure. These challenges are amplified in automatic prompt optimization (\textsc{APO}), where evaluation is repeated throughout the search for better prompts. Studying APO therefore requires a benchmark that is cheap and deterministic to score, hard enough to leave room for improvement, and renewable as models evolve. 
We introduce a chess benchmark built from 1,118 \textsc{Lichess} puzzles to study APO for frozen LLMs: we optimize their prompts without updating their model weights. Chess combines inexpensive exact-match scoring, engine-based evaluation of alternative moves, and a renewable supply of problems with adjustable difficulty. Unlike evaluations that report only success on isolated test items, the benchmark also connects puzzle-solving gains to short game-play rollouts within the same domain.
We use it to evaluate six APO algorithms on eight target models, measuring not only baseline strength but also how much each model responds to optimization and whether optimized prompts transfer across models and to game-play.
Chess is thus a well-suited benchmark for APO: it is (i) \emph{challenging}, as even the strongest evaluated model, Gemini 3.5 Flash (used as the meta-model), solves only about 55\% of puzzles; (ii) \emph{discriminative}, revealing gains, unchanged performance, and regressions across methods and models; (iii) \emph{renewable}, with fresh puzzles to reduce contamination risk and adjustable difficulty to maintain headroom as models improve; and (iv) \emph{affordable}, as the complete study runs for around \$800.
We release the puzzles, optimization and evaluation code, and dataset-renewal scripts (\href{https://github.com/imec-ailabs/Automatic-Prompt-Optimization-with-Chess}{Code}).

\end{abstract}

\section{Introduction}
\label{sec:intro}

Prompt optimization offers a fast, inexpensive way to improve LLM performance on downstream tasks. Automatic prompt optimization (\textsc{APO}) \citep{ramnath2025-apo-survey} refers to methods that automatically search for high-performing prompts, e.g., for instruction generation \cite{zhou2022-ape}.
This reduces the need for costly manual prompt engineering and speeds up the adoption of new models.
Like fine-tuning, APO adapts a trained model to a downstream task, but it only requires black-box access to the model, allowing it to be applied to both open-weight and proprietary models and to be combined with supervised fine-tuning or \textsc{RLHF}.

The main application domains of automatic prompt optimization are text classification, summarization, and reasoning. But evaluating \textsc{APO} inherits every weakness of LLM benchmarking itself: static test sets leak into training data, frontier benchmarks demand heavy infrastructure, grading open-ended outputs requires costly LLM judges, and once the models saturate the task, the benchmarks become obsolete. This motivates benchmarks with cheap, deterministic scoring and hard problems that can be renewed as models evolve.

\begin{figure}[t]
\centering
\begingroup
\definecolor{overviewblue}{RGB}{46,83,112}
\definecolor{overviewgreen}{RGB}{45,119,92}
\definecolor{overviewupdate}{RGB}{170,91,33}
\resizebox{\textwidth}{!}{%
\begin{tikzpicture}[
  x=1mm, y=1mm,
  font=\small,
  box/.style={draw=overviewblue!65, fill=overviewblue!4,
    rounded corners=2pt, line width=.6pt, align=center,
    text width=32mm, minimum height=17mm, inner sep=3pt},
  heading/.style={font=\small\bfseries, text=overviewblue},
  note/.style={font=\footnotesize, align=center},
  arr/.style={-{Latex[length=2mm]}, line width=.8pt, draw=overviewblue},
  lab/.style={font=\footnotesize, fill=white, inner sep=2pt}
]
\node[heading] at (24,0) {(a) Chess puzzle};
\node[heading] at (121,0) {(b) Prompt construction and optimization};
\node[note] at (24,-6) {White to move};
\node[note] at (121,-6) {6 optimizers $\times$ 8 frozen LLMs};

\node[inner sep=0pt] (board) at (24,-32) {%
  \resizebox{48mm}{!}{\chessboard[
    setfen={r1bqkb1r/pppp1ppp/2n2n2/4p2Q/2B1P3/8/PPPP1PPP/RNB1K1NR w KQkq - 4 4},
    boardfontsize=20pt,
    showmover=false,
    color=overviewgreen,
    pgfstyle=straightmove,
    markmoves={h5-f7}
  ]}};
\node[note,anchor=north] at (24,-57) {\textbf{FEN of this board}};
\node[font=\footnotesize\ttfamily,align=left,anchor=north] (fen) at (24,-62)
  {r1bqkb1r/pppp1ppp/2n2n2/4p2Q/\\
   2B1P3/8/PPPP1PPP/RNB1K1NR\\
   w KQkq - 4 4};
\node[note,anchor=north] at (24,-76) {\textbf{Legal moves (UCI, sorted)}};
\node[note,anchor=north] at (24,-81) {\texttt{a2a3}, $\ldots$, \texttt{h5h7}};
\node[note, text=overviewgreen, text width=49mm] (reference) at (24,-94)
  {Reference $y=\texttt{Qxf7\#}$};

\begin{scope}[yshift=8mm]
\draw[draw=overviewblue!65,fill=overviewblue!4,rounded corners=2pt,line width=.6pt]
  (61,-20) rectangle (130,-88);
\fill[overviewupdate!9,rounded corners=2pt] (62,-35) rectangle (129,-60);
\node[font=\small\bfseries,text=overviewblue] at (95.5,-25) {Assembled Prompt};
\node[font=\footnotesize,anchor=west,text=overviewblue] at (64,-31) {JSONAdapter: fixed input/output format};
\node[font=\footnotesize\bfseries,anchor=west,text=overviewupdate] at (64,-39) {Task instructions (part of $p$)};
\node[font=\footnotesize\ttfamily,anchor=west] at (64,-44) {\{task instruction\}};
\draw[overviewupdate!25] (64,-47) -- (127,-47);
\node[font=\footnotesize\bfseries,anchor=west,text=overviewupdate] at (64,-51) {Demonstrations (part of $p$)};
\node[font=\footnotesize\ttfamily,anchor=west] at (64,-56) {\{examples\}};
\node[font=\footnotesize\itshape,anchor=west,text=overviewblue] at (95,-56) {initially empty};
\draw[overviewblue!20] (64,-61) -- (127,-61);
\node[font=\footnotesize\bfseries,anchor=west] at (64,-67) {Current position $x$};
\node[font=\footnotesize\ttfamily,anchor=north west,align=left] at (64,-73)
  {Position Fen: \{fen\}\\
   Side To Move: \{side\}\\
   Legal Moves: \{legal\_moves\}};
\end{scope}

\node[box] (llm) at (164,-67)
  {\textbf{Frozen LLM}\\$\hat y=f(x;\theta,p)$\\[-1pt]
   {\footnotesize weights $\theta$ fixed}};
\node[box] (loss) at (164,-94)
  {\textbf{Loss $\ell(\hat y,y)$}\\{\footnotesize compare with reference\\+ engine evaluation}};
\node[box,draw=overviewupdate,fill=overviewupdate!9,line width=1pt] (optimizer) at (164,-34)
  {\textcolor{overviewupdate}{\textbf{Optimizer}}\\{\footnotesize revise instructions\\and/or demonstrations}};

\draw[arr] (130,-67) -- node[lab,above] {$p,x$} (llm.west);
\draw[arr] (llm.south) -- node[lab,left] {move $\hat y$} (loss.north);
\draw[arr] (loss.east) .. controls +(23,0) and +(23,0) ..
  node[lab,rotate=90,pos=.5] {score / feedback} (optimizer.east);
\draw[-{Latex[length=2.8mm]},draw=overviewupdate,line width=1.5pt]
  (optimizer.west) -- node[lab,above,text=overviewupdate,font=\footnotesize\bfseries]
  {update $p$} (130,-34);

\draw[arr]
  (49,-74) -- node[lab,above] {$x$} (61,-74);
\draw[arr,draw=overviewgreen]
  (reference.east) -- node[lab,above,pos=.9,text=overviewgreen] {$y$} (loss.west);

\end{tikzpicture}%
}
\endgroup
\caption{\textbf{Chess puzzles as a prompt-optimization benchmark.}
\textbf{(a)} A position $x$, its exact FEN (line-wrapped), an abbreviated sorted list of legal moves, and reference move $y$.
\textbf{(b)} The template assembles instructions, demonstrations (initially empty), and the current position fields; braces denote placeholders. APO optimizes the task instructions within the system message and/or demonstrations; the current board and game information are supplied in the user message. The exact template is given in Appendix~\ref{sec:task-prompts}. The warm-colored region and arrow highlight the update of $p$, while $\theta$ stays frozen. Demonstrations include answers, whereas the current reference $y$ is used only for scoring. The optimizer uses training scores and, for reflective methods, textual feedback. The selected $p^*$ is evaluated on held-out puzzles using accuracy and engine-based regret.}
\label{fig:overview}
\end{figure}
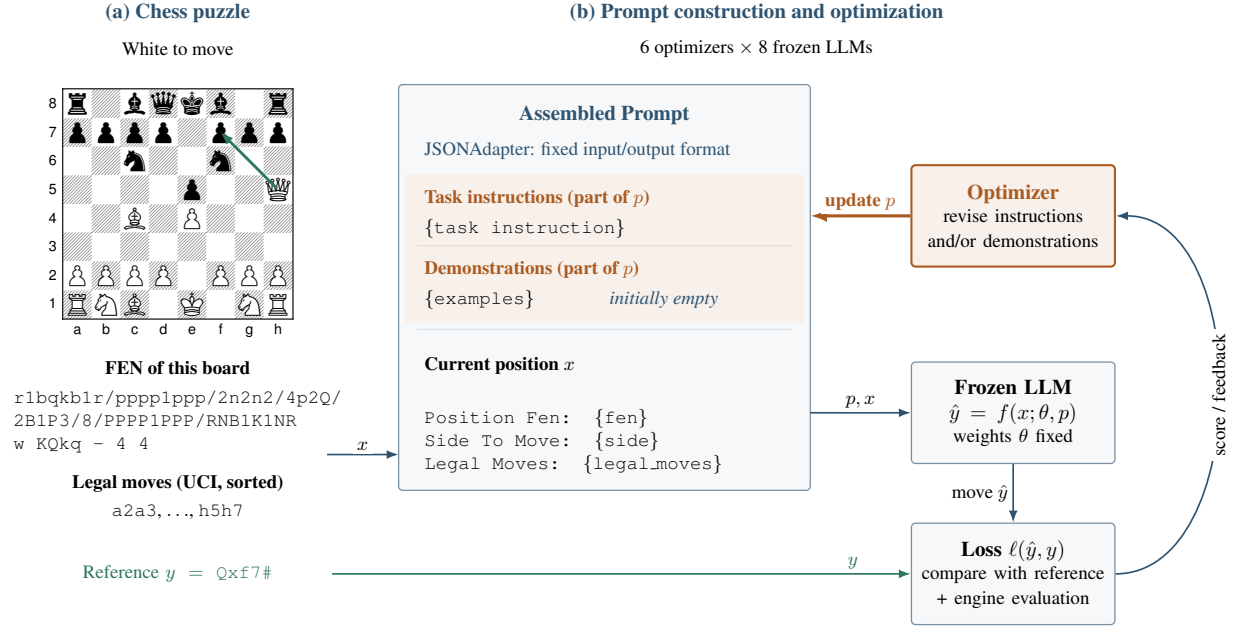

We use chess puzzles to benchmark and understand \textsc{APO}.
Chess has properties that directly address the shortcomings identified above: 
(a) \textit{inexpensive automated scoring}---reference puzzle solutions provide exact-match labels, while engines such as \textsc{Stockfish} \cite{stockfish}, \textsc{Leela Chess Zero} \cite{pascutto2022lc0}, \textsc{Komodo} \cite{Komodo2023}, or \textsc{AlphaZero} \cite{silver2017mastering} evaluate alternative moves without human annotation or an LLM judge;
(b) \textit{renewable problems with tunable difficulty}---the large space of positions supplies challenging problems \citep{kolasani2025-llm-chess}, and scripts can construct fresh puzzle sets from new games and adjust their difficulty as models improve; and
(c) \textit{transfer within the same domain}---prompts optimized on isolated puzzle positions can also be evaluated through their consequences during play.
Puzzles can additionally be generated using RL approaches to target novelty and difficulty \citep{feng2025generating}, at greater computational cost. Figure~\ref{fig:overview} illustrates the benchmark: a frozen LLM receives a chess position as text, and an optimizer updates its instructions and demonstrations using feedback on predicted moves.

\noindent \textbf{Contributions. } 
This paper introduces a new chess benchmark together with an evaluation of six APO algorithms across eight proprietary and open-weight target models. It aims to better understand how prompt optimization behaves across algorithms and models while providing a reproducible way to execute new APO algorithms at a reasonable cost. Our contributions are thus three-fold: (a) a new APO benchmark based on chess puzzles that is easy to run, cheap to evaluate, and challenging yet improvable; (b) a comparison of six APO approaches across eight target models, analyzed along puzzle metadata, prompt composition, engine evaluation, and transferability to other models and game-play rollouts; and (c) an analysis of where optimization gains occur and how they transfer across models and to game-play. In addition, we provide an open-source platform with a curated set of 1,118 puzzles, code to optimize prompts through various standard APO methods, and scripts to refresh the benchmark to reduce dependence on fixed public test sets and adjust difficulty as LLM frontier models improve.

\section{Related Work}
\label{sec:related}

We position our benchmark relative to three lines of work: benchmark construction for frontier LLMs, chess as a testbed for LLMs, and \textsc{APO} algorithms.

\noindent  \textbf{Frontier LLMs Benchmarks.}
APO is typically evaluated on standard LLM benchmarks and inherits their limitations.
As frontier LLMs acquire stronger reasoning capabilities, traditional benchmarks struggle to keep pace and are prone to data saturation and contamination \citep{kiela2021dynabenchrethinkingbenchmarkingnlp,singh2024evaluationdatacontaminationllms,bean2025measuringmattersconstructvalidity,marro2026benchmarkingedgecomprehension}. Moreover, as tasks get more complicated, it becomes harder for humans to produce reliable annotations. \citet{marro2026benchmarkingedgecomprehension} describe this as the ``post-comprehension regime''. 
One response is to generate test problems dynamically instead of fixing a dataset, as in \textsc{LiveBench}~\citep{white2025livebench}, \textsc{RULER} \citep{hsieh2024ruler}, \textsc{LLF-Bench} \citep{cheng2023-llfbench}, and \textsc{LoCoMo} \citep{maharana2024-locomo}. Our benchmark belongs to this renewable-task family. Its puzzle task supports inexpensive exact-match scoring and graded engine evaluation without an LLM judge or an agentic execution environment. This makes repeated prompt evaluation affordable, complementing benchmarks that study open-ended responses with LLM judges \citep{zheng2023-judging,maharana2024-locomo}, agentic tasks in sandboxes \citep{tbench2025-terminal-bench-2,chan2025mle}, or interactive performance in arenas \citep{zheng2023-judging,guertler2025-textarena,liu2025-chessarena}.
\smallskip

\noindent  \textbf{Chess with LLMs.}
Chess has served as a testbed for artificial intelligence for decades, from \textsc{Deep Blue} \citep{campbell2002deepblue} to \textsc{AlphaZero} \cite{silver2018alphazero} and \textsc{Leela Chess Zero} \citep{lc0}.
Existing work uses chess to study LLM capabilities in several ways. Some studies internalize chess behavior by pre-training or heavily fine-tuning model weights on offline game trajectories and state evaluations \citep{noever2020chess,feng2023-chessgpt, zhang2025-chessllm}. In a post-training alternative, \citet{hwang2025-strategic-chess} found that reinforcement learning plateaus at $25\text{--}30\%$ puzzle accuracy, and hypothesized that generic base models lack the foundational state-tracking representations required for RL to build upon. Separately, \citet{karvonen2024-chess-worldmodels} used probes to study internal representations of board-state and latent estimates of opponent skill in models trained in chess transcripts.
Our work instead treats chess as a text-based reasoning task for general-purpose frozen frontier language models. Previous studies show that model performance is sensitive to the representation of the board, such as \textsc{FEN} strings or \textsc{PGN} move lists, and to the prompt structure \citep{acher2023-debunking-chessboard, kolasani2025-llm-chess, cooper2025-pgn2fen}. Manual prompt engineering involving metadata headers, few-shot exemplars, or \textsc{PGN} styles can therefore produce substantial performance differences \citep{dynomight2024-chess-weirdness, zswitten2023-pgn-prompt-style}. Rather than relying on targeted manual interventions, we automate prompt optimization to provide a shared testbed for evaluating Automatic Prompt Optimization (\textsc{APO}) algorithms.

\smallskip

\noindent \textbf{Automatic Prompt Optimization (\textsc{APO}).}
APO adapts LLM behavior without updating the underlying parameters \citep{ramnath2025-apo-survey}. \textsc{APO} approaches typically use LLMs to generate, mutate, and select instructions or demonstrations through discrete search, gradient-free editing, or evolutionary algorithms \citep{jiang2019-lpaqa, zhou2022-ape, prasad2022-grips, pryzant2023-protegi, yang2023-opro}. Frameworks such as \textsc{DSPy} \citep{khattab2024dspy} compile declarative pipelines with optimizers including \textsc{MIPROv2} \citep{opsahlong2024-miprov2}, \textsc{GEPA} \citep{agrawal2025gepa}, and \textsc{SIMBA} \citep{khattab2024dspy} to maximize downstream metrics.

Although existing \textsc{APO} literature highlights gains in tasks like text classification \cite{du2026-tabular-instruction-opt}, question answering \cite{yang2018hotpotqa} , and mathematical reasoning \citep{ramnath2025-apo-survey, cobbe2021gsm8k, aime25}, games offer a complementary setting in which optimized behavior can be evaluated both on isolated decisions and through their consequences during play. Whereas \textsc{Autoharness} \citep{lou2026-autoharness} synthesizes helper code to improve game-play, we isolate the automatic optimization of instructions and demonstrations for frozen LLMs.

\section{Chess Puzzle Benchmark}
\label{sec:tasks}

We use chess to study prompt optimization for general-purpose language models rather than models specifically fine-tuned to play chess games~\citep{lehrach2025codeworldmodelsgeneral, feng2023-chessgpt, zhang2025-chessllm}.

\noindent \textbf{Prompt Optimization with Chess. }
Let $f(x;\theta,p)$ denote the prediction of a model with parameters $\theta$ and prompt $p$ for an input $x$. The loss $\ell(f(x;\theta,p),y)$ measures its error relative to the reference label $y$.
In APO, $f$ is an LLM whose parameters $\theta$ remain frozen. The goal is to find the prompt $p^* = \arg\min_p \sum_{(x,y)} \ell(f(x;\theta,p),y)$. Figure~\ref{fig:overview}(b) illustrates this objective.

The success of APO depends partly on the quality of $\ell$. A useful scoring function should (1) apply to new inputs and predictions, (2) be cheap, since it sits inside the optimization loop, and (3) provide unambiguous feedback, ideally with graded scores. Few benchmarks offer all three properties at once. Open-ended tasks often require costly LLM judgements, while verifiable domains such as math and code typically use binary rewards and require effort to extend their problem sets.

Chess offers all three: an engine can evaluate predicted moves in new positions, cheaply and with graded, reproducible scores. The same engine can supply labels for fresh positions, allowing the benchmark to be renewed to reduce contamination risk and grow in difficulty as models improve.

\noindent \textbf{Chess Puzzle Representation and Evaluation. }
A puzzle comprises $N$ position--move pairs $(x_i,y_i)_{i=0}^{N-1}$ along its reference solution. We query each position independently; the puzzle is solved if $f(x_i;\theta,p)=y_i$ for every $i$. The default loss is the exact-match error; Appendix~\ref{sub:regret} adds a graded engine-based evaluation.

Following~\cite{feng2023-chessgpt,kolasani2025-llm-chess}, the inputs contain the position in the Forsyth--Edwards notation (\textsc{FEN}), the side to move and explicit legal moves~\citep{hwang2025-strategic-chess}.
Although representation preferences vary between models~\citep{kolasani2025-llm-chess}, our preliminary experiments confirmed \textsc{FEN} as an effective input representation for our models.
The model returns one legal move in \textsc{SAN} or \textsc{UCI}, wrapped in JSON. A response that violates this formatting cannot be parsed and is scored as incorrect.
See Figure~\ref{fig:overview} for input construction and Appendices~\ref{sec:chess-notation-examples} and~\ref{sec:example-prompts} for notation and prompt examples.

\noindent \textbf{Dataset Construction and Difficulty. } We construct the benchmark from 1,118 puzzles extracted from the \textsc{Lichess} database, whose exports are released under CC0 1.0 Universal\footnote{\url{https://database.lichess.org/}}. We divide the puzzles equally into training and test sets of 559 puzzles. The puzzles are stratified along two axes: the number of puzzle moves (one to five) and the puzzle elo rating (400 to 2,800). The complete distribution across the length and rating of the puzzle can be found in the Appendix~\ref{app:db_detail}. Because the current benchmark is drawn from the public \textsc{Lichess} database, we cannot exclude pretraining exposure. Its advantage is therefore not guaranteed absence of contamination, but renewability: the same pipeline can construct evaluation sets from newer games and positions.
\textsc{Lichess} ratings estimate difficulty for human solvers and also organize our LLM results: higher-rated puzzles are solved less often (Appendix~\ref{sub:puzzle-difficulty}). In this dataset, one- and two-move puzzles occur below 2000, while three- to five-move puzzles occur above 2000. Rating and length are therefore descriptive axes, not independent difficulty factors or estimates of an LLM's playing rating.

Preliminary experiments showed that recent models do not saturate our puzzle database, making the current set appropriate for APO evaluation.

\section{Experimental Setup}

We compare six \textsc{APO} algorithms across eight target models, using a ninth model, Gemini 3.5 Flash, as the fixed meta-model for prompt proposal and reflection. All target models start from the same instruction: \emph{``Solve the chess puzzle by finding the single best move.''}
The benchmark contains 559 training puzzles and 559 test puzzles. Candidate selection uses the training data according to each optimizer's configuration. We describe the optimizers and feedback below, followed by the model panel and baseline evaluations.

\subsection{Algorithms}
\label{sub:algorithms}

We select six optimizers that cover complementary prompt-improvement strategies: some optimize instructions, some optimize demonstrations, and some jointly optimize both components. They also differ in the type of feedback they exploit and in the search mechanism used to identify improved prompts.

\noindent \textbf{Genetic-Pareto (\textsc{GEPA}) \citep{agrawal2025gepa}:} \textsc{GEPA} employs a genetic-reflective optimization strategy that couples prompt mutation with Pareto-inspired selection. It constructs a Pareto frontier consisting of prompts that perform best on specific subsets of the evaluation set and iteratively evolves the candidates on this frontier. Crucially, \textsc{GEPA} uses execution traces together with textual and numerical feedback to diagnose errors and guide prompt mutations, allowing the search to exploit both qualitative information about model failures and quantitative performance signals.

\noindent \textbf{Cooperative Prompt Optimization (\textsc{COPRO}) \citep{khattab2024dspy}:} \textsc{COPRO} optimizes the instruction component of a prompt through an iterative coordinate-ascent-style search. A meta-model generates multiple candidate instruction variants, which are evaluated using a scalar performance metric. The best-performing candidates are retained and subsequently refined, producing an iterative search over instruction space driven exclusively by numerical feedback.

\noindent \textbf{BootstrapFewShot (\textsc{BFS}) \citep{khattab2024dspy}:} \textsc{BFS} improves prompts by augmenting them with few-shot demonstrations obtained through successful trace bootstrapping. The optimizer executes the program over training examples and retains intermediate execution traces for which the model produces correct outputs. These successful traces are then incorporated into the prompt as demonstrations, making the optimization process dependent on numerical feedback while leaving the instruction component unchanged.

\noindent \textbf{BootstrapRandomSearch (\textsc{BRS}) \citep{khattab2024dspy}:} Like \textsc{BFS}, \textsc{BRS} optimizes prompts by selecting few-shot demonstrations. It  performs a random search over alternative demonstration sets by repeatedly sampling different combinations of demonstrations. It evaluates each configuration on the validation set, and retains those achieving the highest numerical performance. This allows \textsc{BRS} to explore a broader space of possible demonstration configurations than standard bootstrapping.

\noindent \textbf{Multiprompt Instruction Proposal Optimizer Version 2 (\textsc{MIPROv2}) \citep{opsahlong2024-miprov2}:} \textsc{MIPROv2} jointly optimizes prompt instructions and few-shot demonstrations. It first proposes a diverse set of contextualized candidate instructions based on information about the task and program structure, while also bootstrapping examples that the model can solve successfully. It then formulates prompt selection as a joint search problem and uses Bayesian optimization to efficiently explore combinations of candidate instructions and demonstrations. Each configuration is evaluated using a numerical validation metric, and the search progressively focuses on combinations expected to yield higher performance.

\noindent \textbf{Stochastic Introspective Mini-Batch Ascent (\textsc{SIMBA}) \citep{khattab2024dspy}:} \textsc{SIMBA}\footnote{\url{https://blog.mariusvach.com/posts/dspy-simba}} combines few-shot demonstration selection with stochastic introspective optimization of instructions. It samples mini-batches and queries the LLM under different temperature settings to generate diverse execution traces. By comparing successful and unsuccessful traces, as well as identifying examples with high output variability, the optimizer uses self-reflection to derive explicit heuristics or rules that can be incorporated into the prompt. Alternatively, successful traces can be added as demonstrations. In this way, \textsc{SIMBA} exploits both textual feedback derived from introspection and numerical feedback from the evaluation metric while jointly modifying instructions and examples.

\medskip

\begin{table}[!ht]
\centering
\caption{\textbf{Summary of optimization algorithms.} optimization scope (what prompt components are optimized), feedback types (how the algorithm guides the search), and whether a separate meta-model proposes prompt changes.}
\label{tab:algorithm-summary}
\small
\begin{tabular}{lccccc}
\toprule
& \multicolumn{2}{c}{Optimizes} & \multicolumn{2}{c}{Feedback} & \multicolumn{1}{c}{Meta-model} \\
\cmidrule(lr){2-3}\cmidrule(lr){4-5}\cmidrule(lr){6-6}
Algorithm & Instructions & Demonstrations & Textual & Numerical & Used \\
\midrule
\textsc{GEPA} & \checkmark & \checkmark & \checkmark & \checkmark & \checkmark \\
\textsc{COPRO} & \checkmark & & & \checkmark & \checkmark \\
\textsc{BFS} & & \checkmark & & \checkmark & \\
\textsc{BRS} & & \checkmark & & \checkmark & \\
\textsc{MIPROv2} & \checkmark & \checkmark & & \checkmark & \checkmark \\
\textsc{SIMBA} & \checkmark & \checkmark & \checkmark & \checkmark & \checkmark \\
\bottomrule
\end{tabular}
\end{table}

Optimizers receive two types of feedback: textual and scalar.
Textual feedback first classifies failures as parsing errors, illegal outputs, or valid but incorrect answers, and then adds task-specific guidance; when appropriate, the feedback also includes the expected move, \textsc{FEN}, or legal-move set. For example, after a legal but incorrect puzzle move, the feedback states: ``\textit{The move was legal, but did not solve the tactic. Reassess forcing checks, captures, threats, and replies.}''. Non-reflective algorithms only rely on the scalar score, i.e. the puzzle accuracy.
A summary of which algorithm uses which type of feedback can be found in Table~\ref{tab:algorithm-summary}. 

We evaluate each optimizer using its standard recommended configuration. Consequently, the comparison reflects out-of-the-box behavior rather than performance under equalized compute budgets or optimizer-specific hyperparameter tuning.

\subsection{Models and Baseline Evaluations}
\label{sec:model-selection}

Our evaluations span multiple LLM providers: Google, DeepSeek, Anthropic, OpenAI, Alibaba Group and Meta.
We chose the models based on cost, provider diversity and rank in the Kaggle Game Arena~\footnote{\url{https://www.kaggle.com/benchmarks/kaggle/chess-text/versions/2}}.

\begin{table*}[!ht]
\centering
\caption{\textbf{Baseline prompt evaluation on the 559-puzzle test set, reported as mean $\pm$ sample standard deviation over three runs.} Input and output are per solver-position call; cost is normalized per 1,000 solver-position calls. Reasoning: thinking level passed to the model; \emph{default}: provider's default; \emph{N/A}: no reasoning mode. Jev is a decision endpoint, so generative max-output and output-token columns are not comparable.}
\label{tab:model-experiment-characteristics}
\scriptsize
\resizebox{\textwidth}{!}{\begin{tabular}{lrrrrrrrr}
\toprule
Model & Max output & Reasoning & Input/call & Output/call & \$ / 1k calls & Latency/call (s) & Puzzle acc. (\%) & Illegal (\%) \\
\midrule
Gemini 3.5 Flash & 1,024 & medium & $446 \pm 0.00$ & $944 \pm 0.56$ & 9.163 & $5.93 \pm 0.13$ & $55.40 \pm 0.58$ & $0.2 \pm 0.1$ \\
Gemini 3.5 Flash Lite & 1,024 & low & $447 \pm 0.01$ & $14 \pm 0.03$ & 0.169 & $0.67 \pm 0.01$ & $26.30 \pm 0.62$ & $4.9 \pm 1.2$ \\
Claude Haiku 4.5 & 1,024 & default & $646 \pm 0.00$ & $12 \pm 0.00$ & 0.707 & $1.16 \pm 0.02$ & $11.63 \pm 0.00$ & $2.5 \pm 0.0$ \\
GPT-4o Mini & 1,024 & N/A & $427 \pm 0.00$ & $12 \pm 0.03$ & 0.071 & $1.10 \pm 0.03$ & $5.84 \pm 0.37$ & $1.1 \pm 0.1$ \\
Qwen3.8 27B & 512 & default & $452 \pm 0.02$ & $15 \pm 0.01$ & 0.231 & $2.68 \pm 0.84$ & $8.47 \pm 0.37$ & $0.6 \pm 0.3$ \\
GPT-5.6 Luna & 4,096 & low & $426 \pm 0.00$ & $1,529 \pm 2.40$ & 1.920 & $20.87 \pm 1.05$ & $31.72 \pm 0.37$ & $0.5 \pm 0.1$ \\
DeepSeek V4 Pro 0813 & 512 & default & $435 \pm 0.93$ & $13 \pm 0.04$ & 0.559 & $3.10 \pm 0.99$ & $14.49 \pm 0.18$ & $1.3 \pm 0.1$ \\
Muse Spark 1.2 Contributor & 16,384 & low & $424 \pm 0.35$ & $8,227 \pm 44.32$ & 1.677 & $86.17 \pm 3.13$ & $31.90 \pm 1.49$ & $9.0 \pm 1.1$ \\
Jev 1.13 & N/A & N/A & $847 \pm 0.00$ & N/A & 0.036 & $0.32 \pm 0.02$ & $7.27 \pm 0.10$ & $0.0 \pm 0.0$ \\
\bottomrule
\end{tabular}}
\end{table*}

Table~\ref{tab:model-experiment-characteristics} reports the characteristics of the selected models, including price, at the time of the experiments. 
Whenever possible, \verb|reasoning| was set to \textit{low}. We calibrated \verb|max_output_tokens| with preliminary tests to avoid output truncation. Note that Jev 1.13 operates differently from the other models. Instead of generating free text, it acts as a zero-shot classifier that chooses among the legal moves (Appendix~\ref{sec:hps}). It therefore cannot produce illegal moves, and token-based columns do not apply to it.

Table~\ref{tab:model-experiment-characteristics} also reports the baseline results with the base prompt on three seeds in the test set.
Cost and latency vary widely across models.
For example, GPT-5.6 Luna and Muse Spark 1.2 Contributor, while having low nominal prices (input \$0.2 and \$0.1 per M tokens, output \$1.2 and \$0.2 per M tokens, respectively), produce longer outputs under the evaluated configurations, contributing to higher cost and latency. On this benchmark, Gemini 3.5 Flash Lite offers a favorable price-performance trade-off, with 26.30\% accuracy at \$0.169 per 1k calls.
The best model, Gemini 3.5 Flash, is kept as the fixed meta-model for prompt proposal and reflection.
We can see that the illegal move rate varies a lot from one model to another. The list of legal moves is given, so in theory the model should predict only legal moves. However, illegal moves can be due to bad formatting. Tuning the formatting instructions can be a lever on its own to improve performance for some models.

The common baseline instruction is \emph{``Solve the chess puzzle by finding the single best move.''} It is accompanied by the position fields and, for generative models, the JSON adapter's formatting instructions (Appendix~\ref{sec:task-prompts}). We deliberately use a minimal common baseline to study what each optimizer can derive from the same starting point across models, rather than whether APO outperforms the strongest model-specific manual prompts. Despite its simplicity, this baseline is effective for several models. Our focus on affordable models is motivated by evidence that prompt optimization can improve inexpensive, weaker models \citep{khattab2024dspy}, potentially offering an alternative to switching to a more expensive model.

\section{Results}
\label{sec:puzzle-results}
We organize our results around five empirical questions: (1) which models improve, (2) which puzzles account for the gains, (3) what the optimized prompts contain, (4) whether they help other models, and (5) whether puzzle gains extend to short game-play rollouts. 
The eight-model puzzle panel comprises 141 optimization runs and 24 baseline evaluations, with around \$800 in recorded optimization and evaluation costs.

\subsection{Optimization Gains and Evaluation Cost}
\label{sec:puzzle-optim-results}

Table~\ref{tab:puzzle-optim-results} reports the results for the eight models and six optimizers, using Gemini 3.5 Flash as the meta-model.
Detailed optimization configurations are provided in the Appendix~\ref{sec:hps}.
Note that SIMBA is not applicable to Jev 1.13 because it requires changing the temperature to generate outputs, which is not compatible with the 0-shot classification output of Jev.

\begin{table*}[ht]
\centering
\caption{\textbf{Puzzle accuracy (\%, N=559), reported as mean $\pm$ sample standard deviation over three runs per condition.} Bold marks the optimizer with the highest mean for each model. $^{*}$ marks improvement over the model's three-run baseline under an unadjusted one-sided Welch test ($p<0.05$); $^{\dagger}$ marks models with at least one starred optimizer. N.U. marks conditions where none of the three optimizer seeds updated the original prompt. N.A. marks a non-applicable method. Gray cells indicate lower mean accuracy than baseline.}
\label{tab:puzzle-optim-results}
\scriptsize
\setlength{\tabcolsep}{2.5pt}
\resizebox{\textwidth}{!}{\begin{tabular}{lccccccc}
\toprule
Model & Baseline & \textsc{BFS} & \textsc{BRS} & \textsc{COPRO} & \textsc{GEPA} & \textsc{MIPROv2} & \textsc{SIMBA} \\
\midrule
GPT-4o Mini & $5.84 \pm 0.37$ & \cellcolor{gray!20}$5.01 \pm 0.31$ & \cellcolor{gray!20}$4.95 \pm 0.99$ & $5.84 \pm 0.55$ & \cellcolor{gray!20}$5.66 \pm 0.52$ & \cellcolor{gray!20}$4.95 \pm 0.81$ & $\mathbf{5.90 \pm 0.36}$ \\
Jev 1.13$^{\dagger}$ & $7.27 \pm 0.10$ & $7.87 \pm 0.47$ & $8.77 \pm 0.31^{*}$ & $7.69 \pm 0.18^{*}$ & $\mathbf{9.30 \pm 0.47}^{*}$ & $8.59 \pm 0.65^{*}$ & N.A. \\
Qwen3.8 27B$^{\dagger}$ & $8.47 \pm 0.37$ & $9.48 \pm 0.36^{*}$ & $9.78 \pm 0.41^{*}$ & $9.12 \pm 0.31^{*}$ & $\mathbf{10.44 \pm 0.90}^{*}$ & \cellcolor{gray!20}$8.23 \pm 0.31$ & $9.36 \pm 0.45^{*}$ \\
Claude Haiku 4.5 & $11.63 \pm 0.00$ & \cellcolor{gray!20}$11.09 \pm 0.18$ & \cellcolor{gray!20}$11.21 \pm 1.02$ & N.U. & $\mathbf{12.28 \pm 1.22}$ & \cellcolor{gray!20}$10.61 \pm 1.19$ & \cellcolor{gray!20}$11.39 \pm 0.63$ \\
DeepSeek V4 Pro 0813$^{\dagger}$ & $14.49 \pm 0.18$ & $14.55 \pm 0.63$ & $14.85 \pm 1.00$ & $14.79 \pm 0.27$ & $\mathbf{16.28 \pm 0.31}^{*}$ & $14.49 \pm 0.62$ & $15.03 \pm 0.82$ \\
Gemini 3.5 Flash Lite$^{\dagger}$ & $26.30 \pm 0.62$ & $32.38 \pm 0.47^{*}$ & $33.21 \pm 1.15^{*}$ & $29.10 \pm 3.46$ & $27.01 \pm 0.18$ & $32.98 \pm 1.72^{*}$ & $\mathbf{33.87 \pm 1.03}^{*}$ \\
GPT-5.6 Luna & $\mathbf{31.72 \pm 0.37}$ & \cellcolor{gray!20}$23.20 \pm 1.79$ & N.U. & \cellcolor{gray!20}$30.95 \pm 1.46$ & \cellcolor{gray!20}$31.48 \pm 1.17$ & \cellcolor{gray!20}$30.05 \pm 0.78$ & N.U. \\
Muse Spark 1.2 Contributor & $31.90 \pm 1.49$ & \cellcolor{gray!20}$28.92 \pm 1.97$ & \cellcolor{gray!20}$30.59 \pm 0.36$ & N.U. & $32.20 \pm 1.40$ & \cellcolor{gray!20}$29.93 \pm 1.14$ & $\mathbf{32.74 \pm 1.09}$ \\
\bottomrule
\end{tabular}
}
\end{table*}

\begin{figure*}[ht]
    \centering
    \begin{subfigure}{0.495\textwidth}
        \centering
        \includegraphics[width=\textwidth]{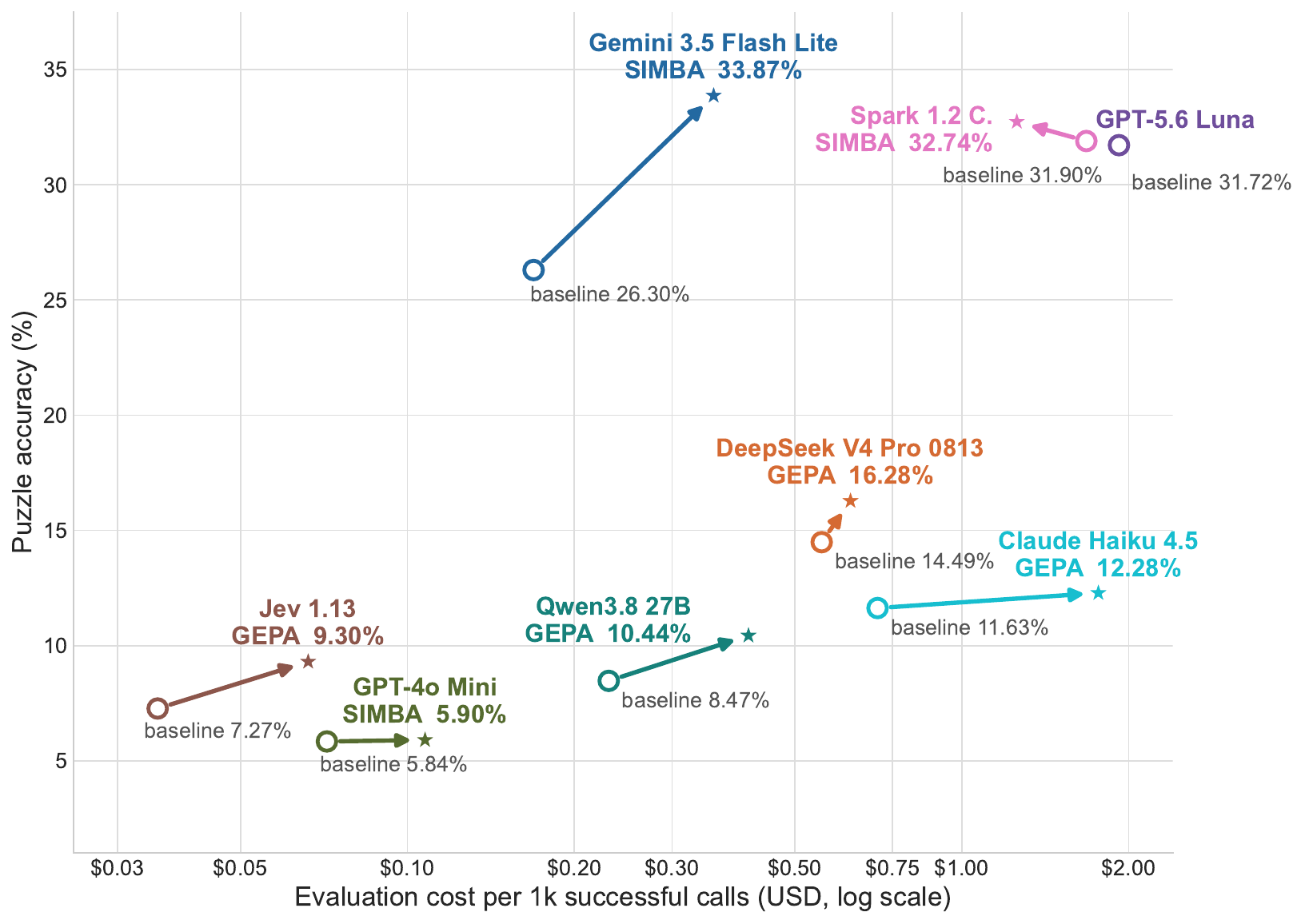}
        \caption{Evaluation cost versus puzzle accuracy. Higher cost is impacted by prompt size and length of reasoning.}
        \label{fig:puzzle-cost-vs-performance}
    \end{subfigure}\hfill
    \begin{subfigure}{0.495\textwidth}
        \centering
        \includegraphics[width=\textwidth]{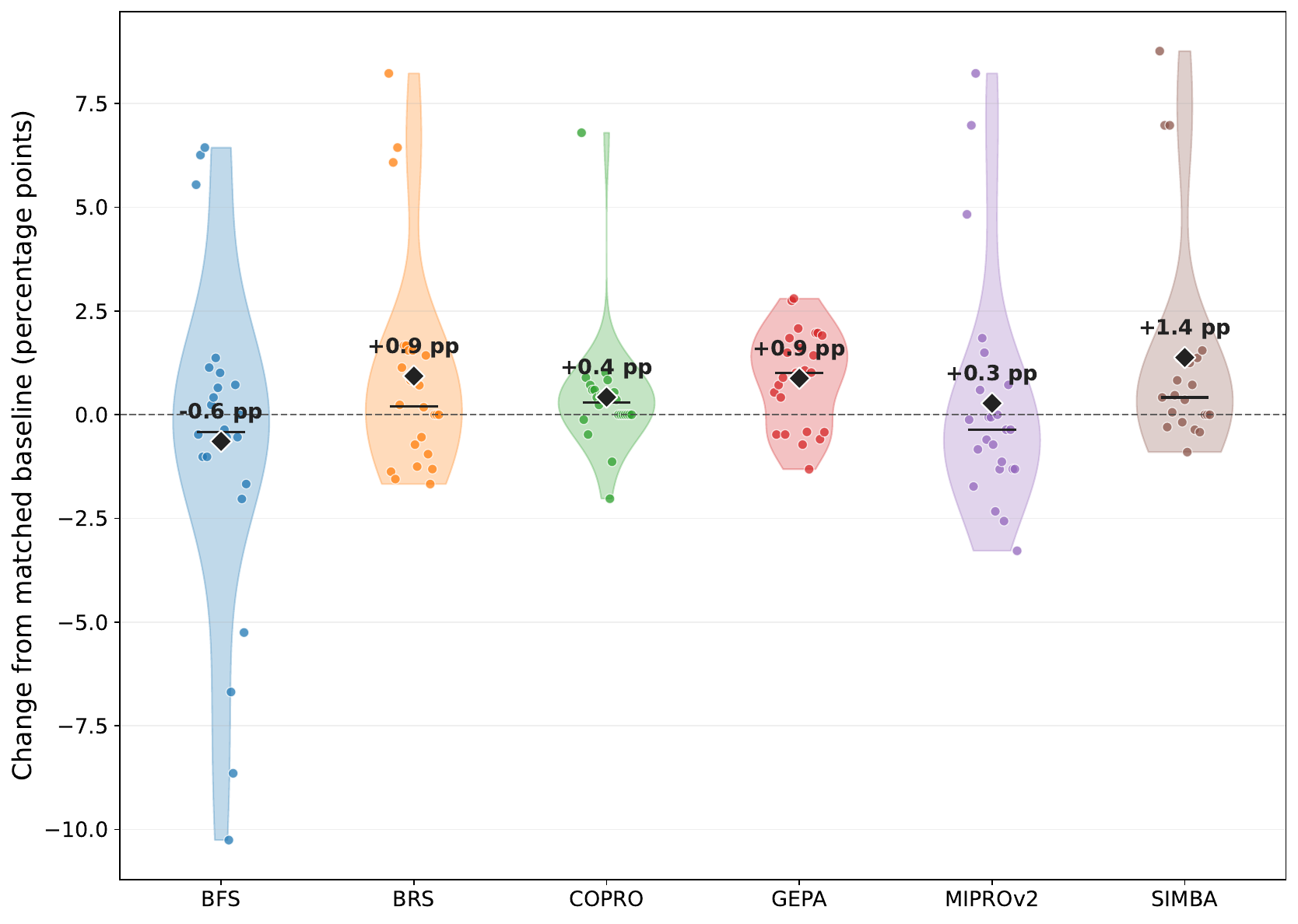}
        \caption{Distribution of Improvement vs Baseline mean in puzzle accuracy by optimizer over all models and seeds.}
        \label{fig:puzzle-algorithm-accuracy-violin}
    \end{subfigure}
    \caption{\textbf{Optimizer performance on the chess puzzle benchmark. }Left: mean evaluation cost versus mean puzzle accuracy across three seeds, from the baseline prompt (open circle) to each model's best algorithm (star). Right: distributions of per-seed changes in puzzle accuracy from the matched baseline for each optimizer.}
\end{figure*}

\noindent \textbf{Observations. } Responsiveness to prompt optimization changes the model ranking: Gemini 3.5 Flash Lite overtakes Luna and Muse despite starting below both (Table~\ref{tab:puzzle-optim-results}). The table reports means and sample standard deviations over three runs; stars mark positive gains over the three-run baseline under a one-sided Welch \(t\)-test (\(p < 0.05\)).
We observe that most models improve with at least one optimizer, but the best optimizer varies from one model to another, as well as the margin of improvement. Conditions marked N.U. are excluded from substantive optimizer interpretation because the optimizer did not find a better prompt in any of the three seeds and returned the original prompt.
The best improvement is observed for Gemini 3.5 Flash Lite: 7.57 percentage points with \textsc{SIMBA}.
Table~\ref{tab:puzzle-optim-results} marks at least one significant gain for Jev 1.13, Qwen3.8 27B, DeepSeek V4 Pro, and Gemini 3.5 Flash Lite. GPT-4o Mini, Claude Haiku 4.5, GPT-5.6 Luna, and Muse Spark show no significant gain. For GPT-5.6, no approaches improved the mean performance compared to baseline. BRS and SIMBA were unable to validate any new candidate prompt and returned the original prompt, resulting in no performance improvement.
The highest mean accuracy is obtained with \textsc{SIMBA} for Flash Lite, GPT-4o Mini, and Muse Spark, and with \textsc{GEPA} for Qwen, DeepSeek V4 Pro, Jev 1.13, and Claude Haiku 4.5. Section~\ref{sub:prompt-transferability} asks whether the resulting prompts remain useful when the target model changes.

Figure~\ref{fig:puzzle-cost-vs-performance} places each model's baseline and best-optimizer accuracy against evaluation cost. Longer instructions and demonstrations increase the input-token burden (Table~\ref{tab:puzzle-prompt-size}), but total cost also depends on the model's output length and pricing. Optimization increases evaluation cost for every model except Spark 1.2 Contributor, which reduces its cost. We hypothesize that, since the model was verbose, prompt optimization managed to channel the reasoning and make it more efficient to answer even if it was not optimized for that.
The separate meta-model, Gemini 3.5 Flash, remains stronger at 55.40\%. Under the baseline prompts, Flash Lite's cost per 1,000 solver-position calls is approximately one tenth that of GPT-5.6 Luna and Muse Spark 1.2 Contributor.

Figure~\ref{fig:puzzle-algorithm-accuracy-violin} shows the performance distribution for each algorithm. 
Although no optimizer wins everywhere, the distributions reveal distinct patterns on this benchmark under the evaluated configurations. \textsc{GEPA} has the highest median gain and improves mean accuracy for 6 out of the 8 models. Unlike \textsc{SIMBA}, it also supports the decision endpoint evaluated here because it does not require a tunable temperature. \textsc{SIMBA} has the largest mean gain and rarely falls below the baseline, but its mean is mainly driven by large improvements on Gemini 3.5 Flash Lite. \textsc{BRS} matches \textsc{GEPA}'s mean gain with a lower median and more spread, and its mean is also mainly driven by gains on Gemini 3.5 Flash Lite. \textsc{COPRO} has the smallest impact on performance, while \textsc{MIPROv2} and \textsc{BFS} fall below baseline more often than not, with \textsc{BFS} showing the largest loss.

\noindent \textbf{Takeaway. } Baseline accuracy does not tell the whole story. Flash Lite starts behind Luna and Muse, yet its larger optimization gain moves it ahead of both. The weakest models, meanwhile, remain far behind despite some significant improvements. This contrast makes responsiveness to optimization a useful comparison axis alongside baseline strength and cost, consistent with earlier evidence that optimizer performance varies across settings \citep{du2026-tabular-instruction-opt, sarmah2024-dspy-teleprompters}.
The benchmark is challenging yet improvable: optimization succeeds in some settings and fails in others, exposing differences between methods and models.
This breadth of comparison is achievable at modest cost.

\subsection{Gains Across Puzzle Difficulty and Structure}
\label{sub:improvement-analysis}

Do optimization gains concentrate in particular puzzles, or are they spread across the benchmark? We examine puzzle ratings and themes to locate the gains, and solver move indices to distinguish improvement on individual decisions from improvement on complete solutions. We focus on the three highest-performing models with significant improvement: Gemini 3.5 Flash Lite, DeepSeek V4 Pro, and Qwen3.8 27B, optimized with \textsc{SIMBA}, \textsc{GEPA}, and \textsc{GEPA}, respectively.

\begin{figure*}[!ht]
    \centering
    \begin{subfigure}{0.495\textwidth}
        \centering
        \includegraphics[width=\textwidth]{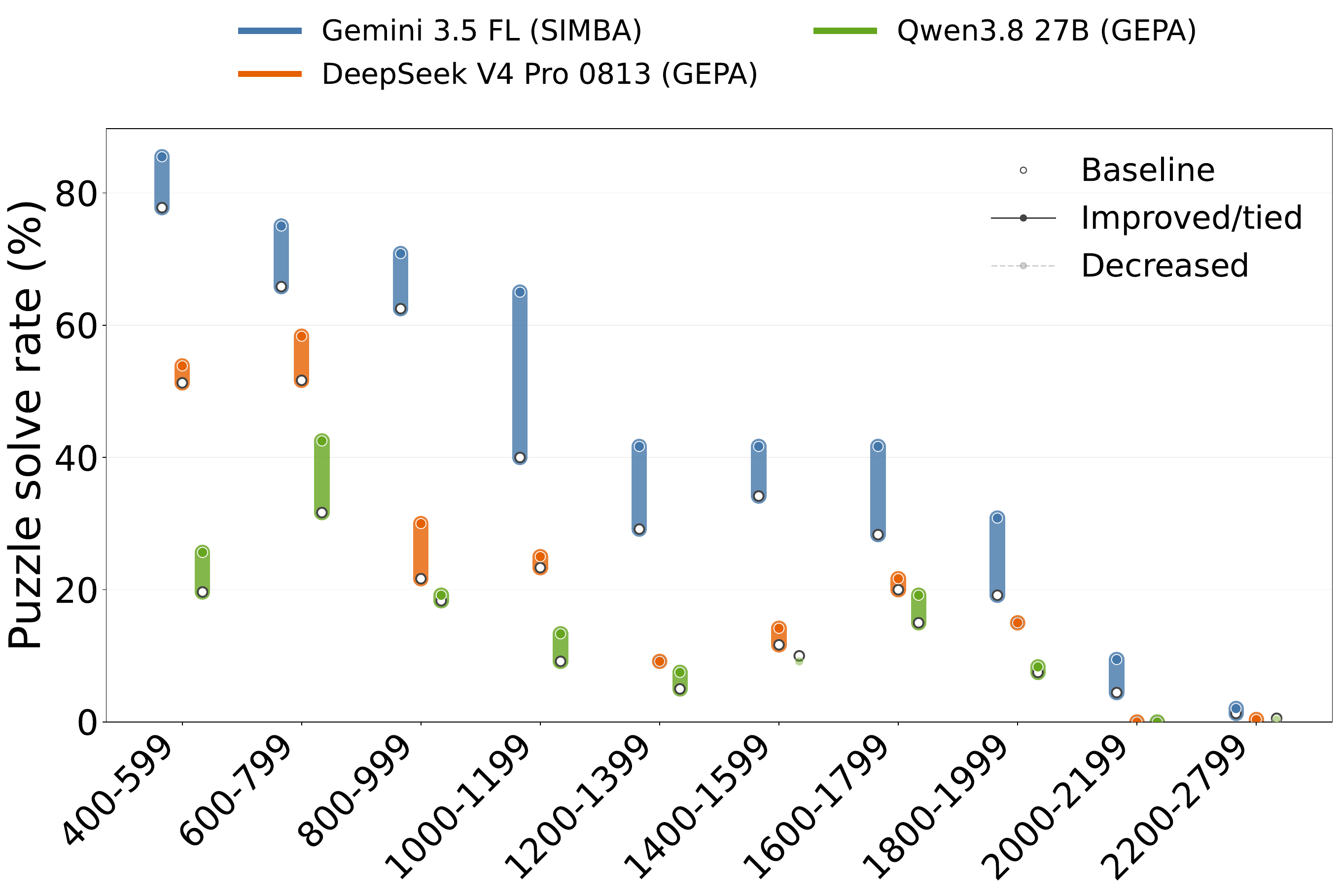}
        \caption{Improvements by \textsc{Lichess} rating.}
        \label{fig:puzzle-baseline-to-best-by-elo}
    \end{subfigure}\hfill
    \begin{subfigure}{0.495\textwidth}
        \centering
        \includegraphics[width=\textwidth]{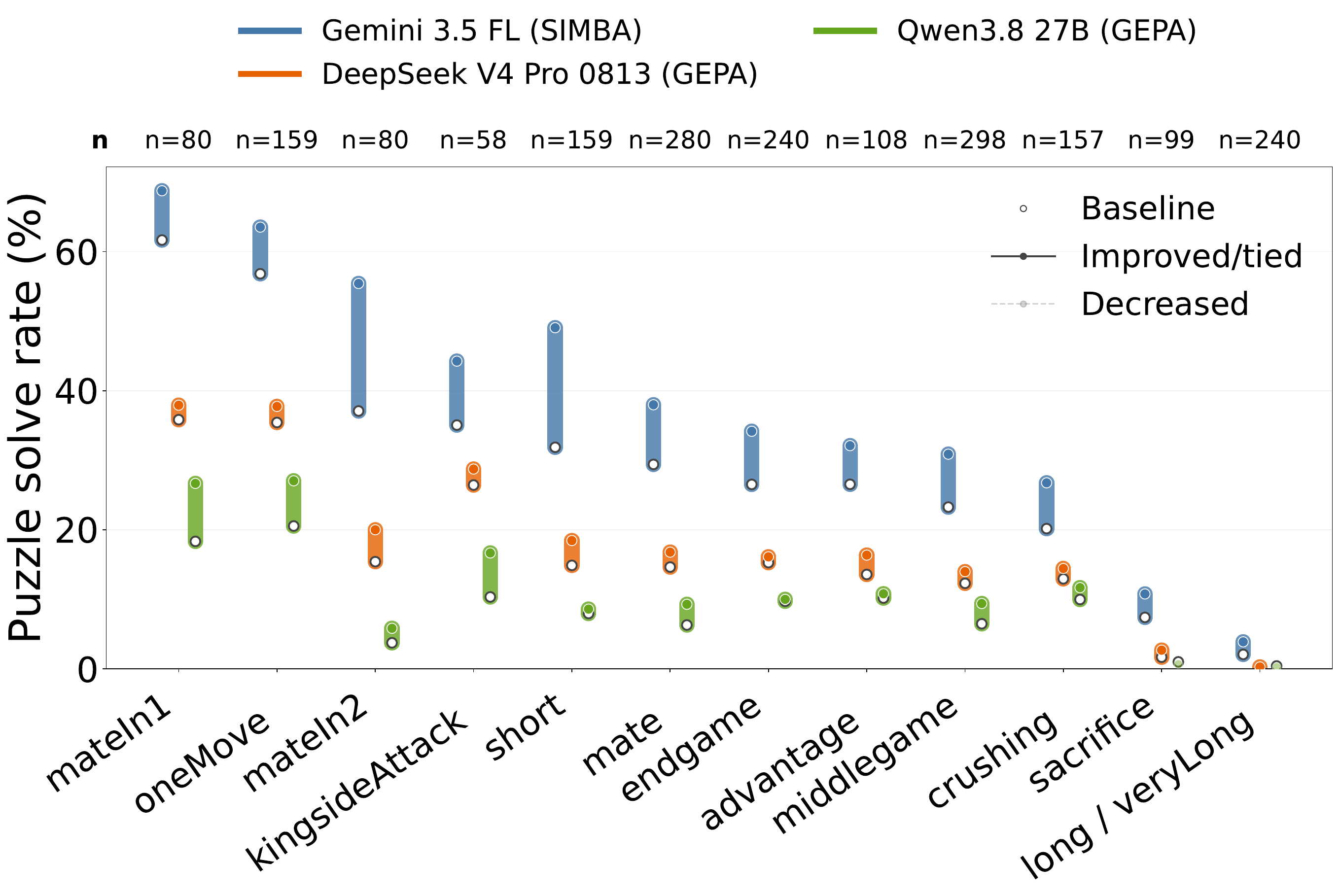}
        \caption{Improvements by Puzzle themes.}
        \label{fig:puzzle-baseline-to-best-by-theme}
    \end{subfigure}
    \caption{\textbf{Puzzle solve rate before and after optimization vs Elo and theme.} We show, for the 3 best models with significant improvement, the results for the best algorithm (cf Table~\ref{tab:puzzle-optim-results}).
    Open circles mark baseline performance and colored circles mark the optimized prompt; faded thinner segments indicate a decrease.}
    \label{fig:puzzle-baseline-to-best-by-slice}
\end{figure*}

\begin{figure*}[!ht]
    \centering
    \includegraphics[width=\textwidth]{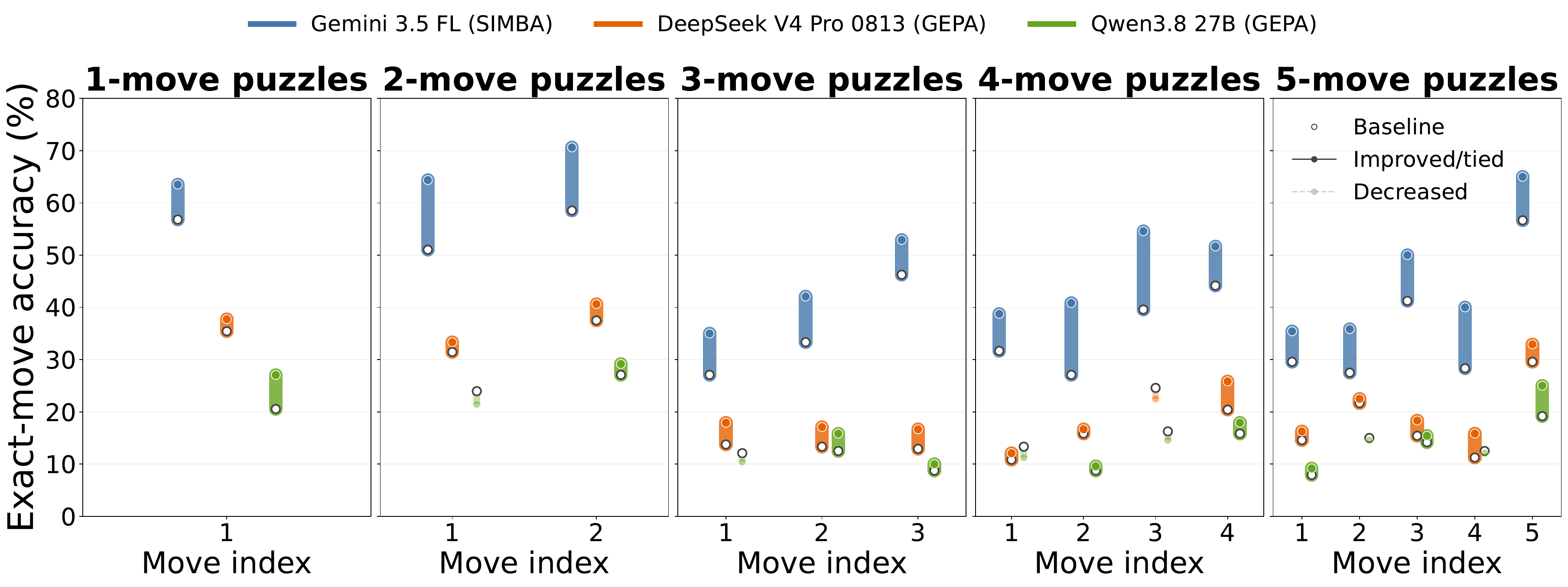}
    \caption{\textbf{Accuracy before and after the best-performing optimizer for each selected model, stratified by puzzle length and solver move index. }Open circles mark baseline performance and colored circles mark the optimized prompt; faded thinner segments indicate a decrease.}
    \label{fig:puzzle-baseline-to-best-by-move-index}
\end{figure*}

\noindent \textbf{Observations. } Figure~\ref{fig:puzzle-baseline-to-best-by-slice} decomposes the improvement by Elo range and theme.
In Figure~\ref{fig:puzzle-baseline-to-best-by-elo}, we see that the improvement spreads across a large Elo range but largely stays below 2200 Elo, with only Gemini 3.5 Flash Lite improving on puzzles between 2000 and 2200 Elo.
The gains are not uniform: some bins are left unchanged or slightly penalized by the new prompt. For example, Qwen3.8 27B loses 0.8 points on puzzles between 1400 and 1600 Elo, and DeepSeek V4 Pro stays exactly flat between 1200 and 1400 Elo.
Figure~\ref{fig:puzzle-baseline-to-best-by-theme} plots the improvement on the 12 most frequent themes in the database. Aggregated over the three models, the mateIn2 and short themes capture the most improvement, followed by kingsideAttack, mateIn1, oneMove, and mate. The ranking varies per model, for example Qwen3.8 27B improves most on mateIn1 and oneMove puzzles.
Note that puzzles have several themes, so improvement on one puzzle may impact multiple themes, for example mate, mateIn1, and mateIn2 may have large overlap.
Figure~\ref{fig:puzzle-baseline-to-best-by-move-index} shows the improvement at each individual move index of the puzzles.
The 3+ move puzzles cover the 2000+ Elo range. Despite the low full-puzzle improvement in this range, per-move accuracy still improves at most move indexes for all models.
We still observe some exceptions, with Qwen3.8 27B decreasing on the first move of two-move puzzles and on the third move of four-move puzzles. A significant overall improvement can therefore hide local losses.
We also observe that the last move of a puzzle is often the most accurately solved, most clearly in five-move puzzles. Early moves require anticipating the entire solution line, while the last move concludes the tactic, so uncertainty is greatest at the start of a puzzle.

\noindent \textbf{Takeaway. } Observed gains span several themes and rating ranges, predominantly below a puzzle rating of 2200. 
A higher overall score can hide both newly solved puzzles and lost capabilities on particular subsets. 
Rating, theme, and move-index views help locate these trade-offs.
Longer puzzles pose a particular challenge: improving individual moves is not enough unless those gains combine into a fully correct solution. 
Their low solve rates leave room for methods that improve consistency across the entire reference line.

The gains span several puzzle categories. We next examine what the optimizers changed in the prompts.

\subsection{What Do Optimized Prompts Contain?}
\label{sub:prompt-content-analysis}

\begin{figure*}[!ht]
    \centering
    \includegraphics[width=\textwidth]{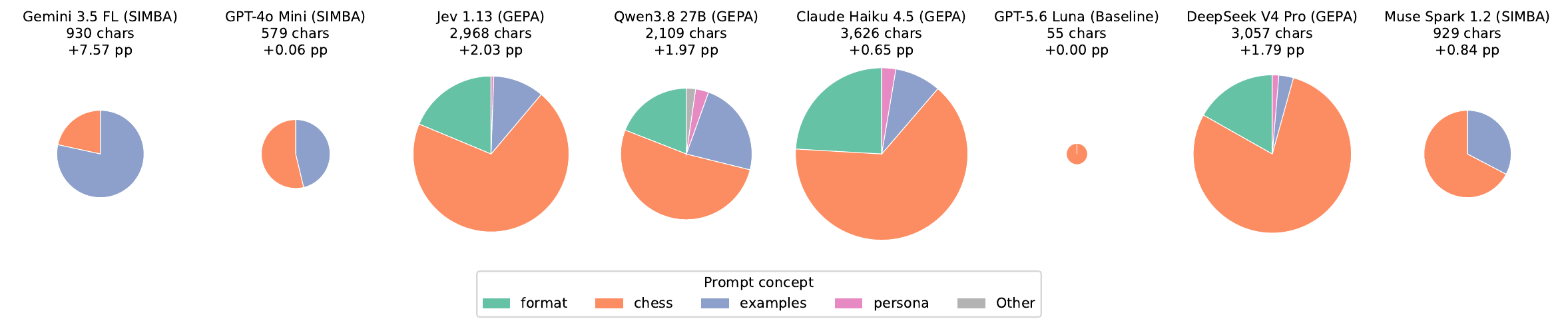}
    \caption{\textbf{Mean composition of the three prompts produced by each model's best algorithm.} Wedge proportions represent each category's share of prompt characters; pie area represents total prompt length in characters.}
    \label{fig:best-prompt-composition-by-model}
\end{figure*}

To better understand the impact of optimized prompts on performance, we analyze the prompt composition of the best algorithms.
We classify best prompt content into four categories: (1) \emph{formatting instructions} that constrain the output structure, (2) \emph{chess instructions} that give chess-specific guidance such as checking forcing moves first, (3) \emph{examples}, including illustrative examples and demonstrations and (4) \emph{persona framing} that assigns the model a role. We report the average composition over 3 seeds for each model in Figure~\ref{fig:best-prompt-composition-by-model}.
Instruction text is split into sentence-like units, treating list items as individual units, and labeled with Gemini 3.1 Flash Lite at temperature 0 without reasoning. The classifier receives all instruction units from a compiled prompt in one call; stored demonstrations are assigned directly to the examples category. 
We map the classifier's unclassified output to Other. 
Character counts in categories preserve the full instruction text, including whitespace, and the displayed demonstration fields.

\noindent \textbf{Observations. } 
We see in Figure~\ref{fig:best-prompt-composition-by-model} that prompts produced by the same algorithm have similar compositions across models: \textsc{SIMBA} emphasizes chess guidance and examples, whereas \textsc{GEPA} produces longer chess instructions together with formatting and persona framing. 
Flash Lite achieves the largest gain with prompts averaging just 930 characters: about 78\% examples and 22\% chess guidance, with no additional formatting instructions. The selected GEPA prompts average 2,109--3,626 characters, showing that the largest gain does not require the longest prompt.
Other optimizers are not present in the plot but we know from Table~\ref{tab:algorithm-summary} that \textsc{COPRO} is instruction-only, while \textsc{BRS} and \textsc{BFS} add only examples to the baseline prompt. Therefore they have less diversity. \textsc{MIPROv2} can combine examples and instructions, but it did not produce the best prompt for any model.
Additional details and examples can be found in the Appendix~\ref{sec:prompts-size}.

\noindent \textbf{Takeaway. } APO algorithms significantly influence the composition of the prompts. The algorithms that can combine several content types, \textsc{SIMBA} and \textsc{GEPA}, produce the best results for seven of the eight models. The best prompts do not simply include more of everything: the largest gain comes from a prompt with only two categories, and no winning prompt uses all the content types.

These different prompt compositions raise a further question: are they useful task guidance or adaptations to a particular model?

\subsection{Are Optimized Prompts Generic or Model Specific?}
\label{sub:prompt-transferability}

Does APO discover general task guidance, or does it adapt prompts to a particular model? We investigate this question by applying each selected prompt unchanged to other models. Keeping the puzzle set fixed lets us examine generalization across models rather than across puzzles. Gains on other models would indicate that the prompt's usefulness extends beyond its source, while uneven or negative transfer would reveal dependence on the recipient.
For each of seven source models, we select the prompt with the highest single-run test-set puzzle accuracy. This differs from the three-run optimizer ranking in Table~\ref{tab:puzzle-optim-results}: for example, GPT-4o Mini contributes a \textsc{COPRO} prompt even though \textsc{SIMBA} has the higher mean accuracy. GPT-5.6 Luna remains a target but not a source because its selected prompt is unchanged from the baseline.

\begin{figure*}[!ht]
    \centering
    \includegraphics[width=0.92\textwidth]{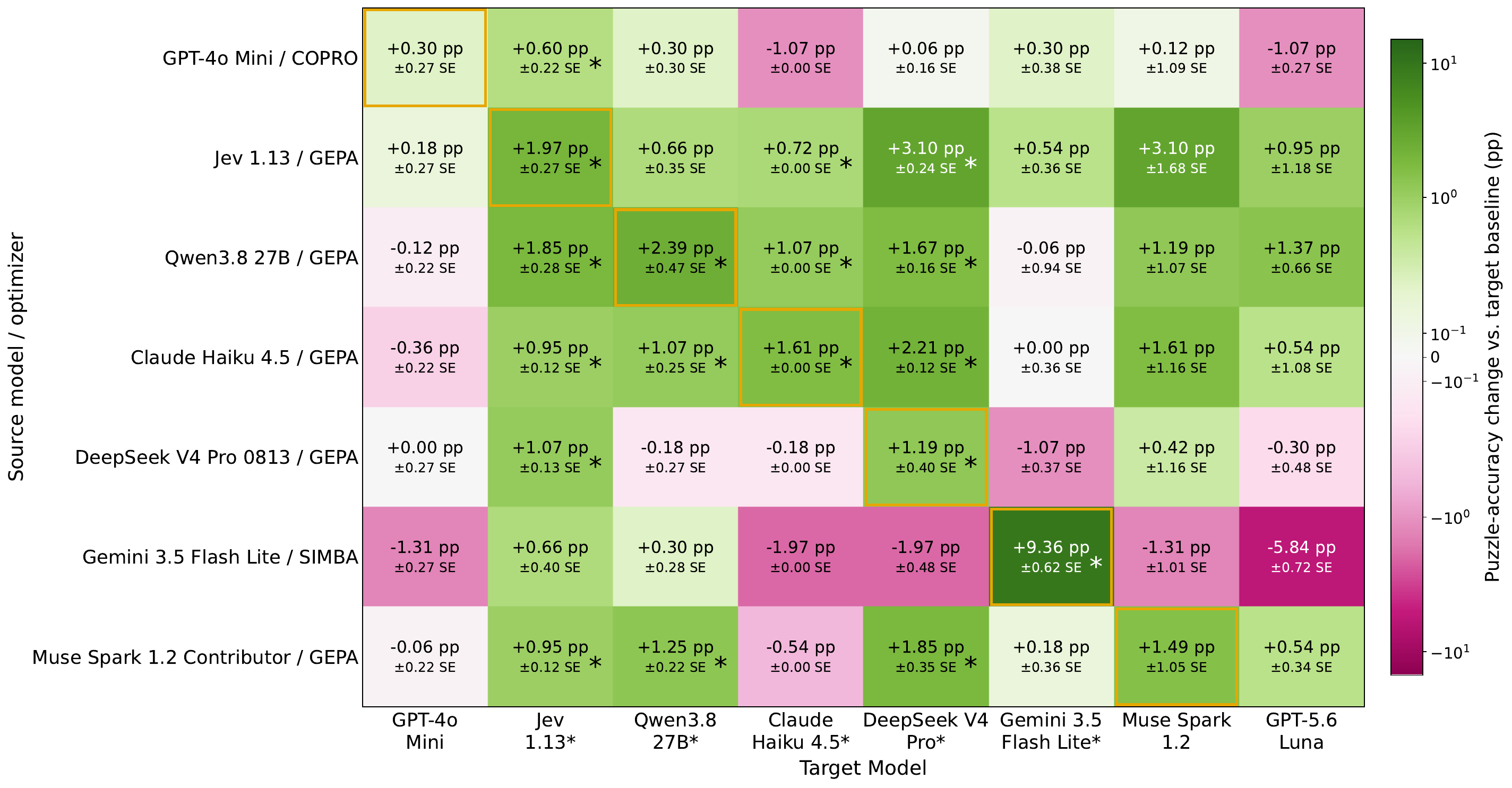}
    \caption{\textbf{Prompt transferability on 559 puzzles.} Rows identify the source model and optimizer; columns identify the target model. Cells report mean puzzle-accuracy change in percentage points relative to the target's three-run baseline, with $\pm$ one standard error of the difference, including variability in both groups. Cell stars mark positive gains with one-sided Welch-test $p<0.05$. A target-model star indicates at least one starred cell in its column, including the native prompt. Gold outlines mark native prompts.}
    \label{fig:prompt-transferability}
\end{figure*}

\noindent \textbf{Observations. } Each populated cell in Figure~\ref{fig:prompt-transferability} compares three evaluations of the same selected prompt with the three baseline cohorts run on three dates summarized in Table~\ref{tab:model-experiment-characteristics}. We computed the differences in the group means and used one-sided Welch tests.
\footnote{Claude Haiku 4.5 returned identical outputs in all seeds, despite caching deactivated, so we mark nonzero differences as significant.}
Across the 49 evaluated cross-model pairs, 31 have positive mean changes, 16 have negative changes, and two are unchanged. Transfer therefore depends on the source--target pairing rather than providing a consistent benefit.
Qwen3.8 27B, Jev 1.13, Claude Haiku 4.5, and DeepSeek V4 Pro receive significant improvements from prompts optimized for other models.
Despite its different serving interface, Jev 1.13 is the only source whose prompt improves all seven other models, with mean changes from +0.18 to +3.10 points. Its prompt raises DeepSeek V4 Pro's accuracy from 14.49\% to 17.59\%, showing that useful guidance can transfer from a weaker source model to a stronger recipient. By contrast, the prompt with the largest native gain, selected for Gemini 3.5 Flash Lite, degrades five of the seven other models.

\noindent \textbf{Takeaway. } Native improvement and portability are distinct: Jev's prompt improves all seven other models, whereas Flash Lite's prompt, which has the largest native gain, degrades five of them. Optimized guidance can extend beyond its source model, but its usefulness depends on the recipient.
This matches the literature: cross-model transfer of optimized prompts has been known since \citet{jiang2019-lpaqa}, with prompts from small models transferring to larger ones while the reverse degrades \citep{deng2022-rlprompt}. \citet{agrawal2025gepa} even report small-model prompts beating optimizers running directly on the larger target. In our experiments this pattern appears only in a few cases, for example the Jev 1.13 prompt brings DeepSeek V4 Pro above its own best optimized accuracy.

Cross-model transfer changes the model while holding the puzzle task fixed. We now change the setting: do puzzle-optimized prompts also help when the model plays a sequence of moves against an opponent?

\subsection{Transfer from Puzzles to Play}
\label{sub:playing-chess-transfer}

Chess play requires tactics as well as decision making in balanced positions. We therefore evaluate whether prompts selected by puzzle accuracy transfer to game-play rollouts against \textsc{Stockfish} (GPL-3.0 License\footnote{\url{https://github.com/official-stockfish/Stockfish?tab=GPL-3.0-1-ov-file}}) at depth 20.
We evaluate ten starting conditions: the initial board with the model playing as White or Black, and eight positions from common openings (details in Appendix~\ref{sec:playing-chess-starting-positions}). For each rollout, the model predicts at most ten model moves.
As for puzzles, model move predictions are independent calls that receive only the prompt, the current \textsc{FEN}, color to play, and legal moves; \textsc{Stockfish} replies after every accepted move. We compare each model's base prompt with its best changed held-out puzzle prompt; Gemini 3.5 Flash is shown only with its base prompt.

Unlike exact-match puzzle scoring, game-play evaluation does not require a unique reference move.
At each model turn, regret is the difference between \textsc{Stockfish}'s depth-20 evaluations of its best move and of the model's move. 
Regret is measured only for accepted moves; illegal or unparseable attempts terminate the rollout.
Rollouts also stop at ten model moves, a terminal board state, or a timeout. 
Because prompts induce different trajectories and termination patterns, regret compares move quality on each prompt's own visited states, conditional on producing accepted moves, rather than on a shared position set.
The preliminary evaluation of the relationship between regret evaluation in puzzles vs binary evaluation in puzzles can be found in the Appendix~\ref{sub:regret}. Those results showed that both metrics lead to a similar ranking of the best models and prompts.

\begin{table*}[!ht]
\centering
\caption{\textbf{\textsc{Stockfish} depth-20 regret per accepted model move.} For the overall mean, accepted moves are pooled across starting positions within each seed, then the three seed means are summarized as mean $\pm$ sample SD. Thus seeds have equal weight, but longer rollouts contribute more within a seed. Median regret pools accepted moves across seeds. Regret values are in centipawns; lower is better. Illegal-attempt rates pool flagged-illegal attempts, including unparseable outputs, over all recorded model attempts across seeds; counts are shown below. Checkmate-loss rates count rollouts ending with the model checkmated over all rollouts. Timeouts are not counted as additional illegal attempts. Base and Best give rates before and after prompt optimization. Bold marks the lower value within each base/best model pair, without implying significance.}
\label{tab:playing-chess-transfer}
\scriptsize
\setlength{\tabcolsep}{3.5pt}
\resizebox{\textwidth}{!}{\begin{tabular}{l|r|rr|rr|rr}
\toprule
& \multicolumn{1}{c}{Gemini 3.5 Flash} & \multicolumn{2}{c}{Gemini 3.5 Flash Lite} & \multicolumn{2}{c}{DeepSeek V4 Pro} & \multicolumn{2}{c}{Qwen3.8 27B} \\
\cmidrule(lr){2-2}\cmidrule(lr){3-4}\cmidrule(lr){5-6}\cmidrule(lr){7-8}
Starting position & Base & Base & Best & Base & Best & Base & Best \\
\midrule
Initial, White & 7.78 & 101.00 & \textbf{49.71} & 3,617.25 & \textbf{3,481.93} & \textbf{9,993.60} & 10,001.30 \\
Initial, Black & 1.90 & \textbf{13.50} & 14.64 & \textbf{52.81} & 61.37 & \textbf{7,717.38} & 10,698.79 \\
Ruy Lopez & 6.50 & 133.23 & \textbf{84.57} & 3,383.90 & \textbf{72.50} & \textbf{89.60} & 3,390.57 \\
Open Sicilian & 18.10 & 82.29 & \textbf{69.37} & \textbf{3,566.48} & 6,937.93 & 10,001.10 & \textbf{9,995.70} \\
Queen's Gambit & 24.63 & 48.43 & \textbf{25.04} & 13,627.50 & \textbf{5,938.82} & 10,065.25 & \textbf{9,993.60} \\
King's Indian & 23.20 & \textbf{68.50} & 69.85 & \textbf{3,670.04} & 11,537.35 & \textbf{7,196.93} & 9,998.20 \\
French Defence & 14.10 & 55.27 & \textbf{38.19} & \textbf{9.00} & 4,102.76 & 6,911.52 & \textbf{3,641.25} \\
Caro--Kann & 0.40 & \textbf{41.23} & 58.30 & 13,323.20 & \textbf{9,980.97} & 9,148.95 & \textbf{4,620.77} \\
English Opening & 25.57 & \textbf{24.30} & 44.17 & 3,395.70 & \textbf{110.27} & 112.60 & \textbf{111.00} \\
Nimzo--Indian & 55.70 & \textbf{45.77} & 56.03 & 3,401.50 & \textbf{3,361.57} & 9,992.00 & \textbf{7,441.85} \\
\midrule
Overall mean & $17.89 \pm 2.49$ & $56.01 \pm 1.85$ & $\mathbf{52.65 \pm 4.97}$ & $4,608.22 \pm 738.93$ & $\mathbf{4,517.15 \pm 1,061.59}$ & $7,094.24 \pm 2,722.68$ & $\mathbf{7,030.45 \pm 997.09}$ \\
Overall median & 3 & \textbf{8} & \textbf{8} & 47 & \textbf{33} & 71 & \textbf{48} \\
\midrule
Illegal / total attempts & 3/300 & \textbf{5/278} & 10/267 & 5/247 & \textbf{3/271} & \textbf{0/275} & 2/287 \\
Checkmate losses / games & 0/30 & 0/30 & 0/30 & 9/30 & \textbf{6/30} & 14/30 & \textbf{9/30} \\
\bottomrule
\end{tabular}}
\end{table*}

\noindent \textbf{Observations. } Puzzle gains are not accompanied by a statistically significant reduction in mean regret in the evaluated rollouts. Table~\ref{tab:playing-chess-transfer} reports the same three models as Section~\ref{sub:improvement-analysis}, together with the meta-model's baseline. Mean regret decreases slightly for all three optimized models; median regret and checkmate-loss counts decrease or remain unchanged, while illegal-attempt rates increase for Flash Lite and Qwen and decrease for DeepSeek.
Note that the denominators of the illegal-attempt rates vary because rollout length depends on illegal attempts and checkmates.
Legal moves alone do not ensure good play: Qwen rarely makes illegal moves with its baseline prompt, yet is frequently checkmated.

\noindent \textbf{Takeaway. } Puzzle gains did not produce a statistically significant reduction in mean regret over accepted moves in the evaluated rollouts. Better tactical puzzle solving does not, by itself, establish better play from standard opening positions.
This game-level evaluation is valuable beyond the transfer question itself. Because the same engine-based regret scores both puzzle solving and game-play rollouts, we can directly estimate the gap between benchmark gains and performance in this broader setting.
Chess thus highlights how much an improvement in one specific skill contributes to the overall task, whereas in most domains, coding, mathematics, or question answering, this benchmark-to-deployment gap is far harder to estimate.

\section{Conclusion}

We introduced a chess-puzzle benchmark and compared six prompt optimizers across eight target models. Four models show significant puzzle-accuracy gains, with improvements reaching 7.57 percentage points, while the hardest puzzles remain largely out of reach. The study distinguishes three properties: baseline strength, responsiveness to optimization, and transferability.

Stronger models do not necessarily improve more, and larger gains do not guarantee better transfer. Some prompts help other models, but we find no clear reduction in mean regret in the evaluated game-play rollouts. The benchmark reveals not just how much prompts improve, but where those improvements apply.

Chess makes repeated evaluation inexpensive, datasets renewable, and difficulty adjustable as models improve. It also lets us test whether puzzle gains carry over to play. Together, these properties make it an affordable testbed for APO before evaluation in more costly application domains.

\bibliography{main}
\bibliographystyle{unsrtnat}

\clearpage
\appendix

\section*{Appendix}
\label{sec:appendix}

\section{Chess Details}

\subsection{Chess Notation Examples}
\label{sec:chess-notation-examples}

 Forsyth--Edwards Notation (\textsc{FEN}) describes one board state, including the side to move, castling rights, en-passant target, and move counters.
After the moves \texttt{1. e4 e5 2. Nf3}, the current position is represented in \textsc{FEN} as
\begin{center}
\small\texttt{rnbqkbnr/pppp1ppp/8/4p3/4P3/5N2/PPPP1PPP/RNBQKB1R b KQkq - 1 2}.
\end{center}
 Universal Chess Interface (\textsc{UCI}) notation represents the same move sequence using source and destination squares: \texttt{e2e4 e7e5 g1f3}. Portable Game Notation (\textsc{PGN}) records the game sequence in standard chess notation: \texttt{1. e4 e5 2. Nf3}.

\subsection{Playing Chess Starting Positions}
\label{sec:playing-chess-starting-positions}

\begin{table*}[!ht]
\centering
\caption{\textbf{Starting positions used for the playing-chess transfer evaluation.} ``Model color'' determines which side is controlled by the language model.
}
\label{tab:playing-chess-starting-positions}
\scriptsize
\setlength{\tabcolsep}{4pt}
\begin{tabular}{lll}
\toprule
Position & Model color & Starting \textsc{FEN} \\
\midrule
Initial, White & White & \texttt{rnbqkbnr/pppppppp/8/8/8/8/PPPPPPPP/RNBQKBNR w KQkq - 0 1} \\
Initial, Black & Black & \texttt{rnbqkbnr/pppppppp/8/8/8/8/PPPPPPPP/RNBQKBNR w KQkq - 0 1} \\
Ruy Lopez & White & \texttt{r1bqkbnr/1ppp1ppp/p1n5/1B2p3/4P3/5N2/PPPP1PPP/RNBQK2R w KQkq - 0 4} \\
Open Sicilian & White & \texttt{rnbqkbnr/pp2pppp/3p4/8/3pP3/5N2/PPP2PPP/RNBQKB1R w KQkq - 0 4} \\
Queen's Gambit & White & \texttt{rnbqkb1r/ppp2ppp/4pn2/3p4/2PP4/2N5/PP2PPPP/R1BQKBNR w KQkq - 2 4} \\
King's Indian & White & \texttt{rnbqk2r/ppppppbp/5np1/8/2PP4/2N5/PP2PPPP/R1BQKBNR w KQkq - 2 4} \\
French Defence & Black & \texttt{rnbqkbnr/ppp2ppp/4p3/3p4/3PP3/2N5/PPP2PPP/R1BQKBNR b KQkq - 1 3} \\
Caro--Kann & Black & \texttt{rnbqkbnr/pp2pppp/2p5/3p4/3PP3/2N5/PPP2PPP/R1BQKBNR b KQkq - 1 3} \\
English Opening & Black & \texttt{rnbqkb1r/pppp1ppp/5n2/4p3/2P5/2N3P1/PP1PPP1P/R1BQKBNR b KQkq - 0 3} \\
Nimzo--Indian & Black & \texttt{rnbqk2r/pppp1ppp/4pn2/8/1bPP4/P1N5/1P2PPPP/R1BQKBNR b KQkq - 0 4} \\
\bottomrule
\end{tabular}
\end{table*}

\subsection{Puzzle Dataset Details}
\label{app:db_detail}

\begin{table}[h]
\centering
\caption{\textbf{Number of training/test puzzles by solver moves and \textsc{Lichess} rating bin.} Each cell is reported as training/test.}
\label{tab:test-set-distribution}
\scriptsize
\resizebox{\textwidth}{!}{\begin{tabular}{lcccccccccccc|c}
\toprule
 & \multicolumn{8}{c}{Rating 400--1999} & \multicolumn{4}{c|}{Rating 2000--2799} & \\
Moves & \rotatebox{70}{400} & \rotatebox{70}{600} & \rotatebox{70}{800} & \rotatebox{70}{1000} & \rotatebox{70}{1200} & \rotatebox{70}{1400} & \rotatebox{70}{1600} & \rotatebox{70}{1800} & \rotatebox{70}{2000} & \rotatebox{70}{2200} & \rotatebox{70}{2400} & \rotatebox{70}{2600} & Total \\
\midrule
1 & 19/\textbf{19} & 20/\textbf{20} & 20/\textbf{20} & 20/\textbf{20} & 20/\textbf{20} & 20/\textbf{20} & 20/\textbf{20} & 20/\textbf{20} & -- & -- & -- & -- & 159/\textbf{159} \\
2 & 19/\textbf{20} & 20/\textbf{20} & 21/\textbf{20} & 20/\textbf{20} & 20/\textbf{20} & 20/\textbf{20} & 20/\textbf{20} & 20/\textbf{20} & -- & -- & -- & -- & 160/\textbf{160} \\
3 & -- & -- & -- & -- & -- & -- & -- & -- & 20/\textbf{20} & 20/\textbf{20} & 19/\textbf{20} & 21/\textbf{20} & 80/\textbf{80} \\
4 & -- & -- & -- & -- & -- & -- & -- & -- & 20/\textbf{20} & 20/\textbf{20} & 20/\textbf{20} & 20/\textbf{20} & 80/\textbf{80} \\
5 & -- & -- & -- & -- & -- & -- & -- & -- & 19/\textbf{20} & 21/\textbf{20} & 20/\textbf{19} & 20/\textbf{21} & 80/\textbf{80} \\
\midrule
Total & 38/\textbf{39} & 40/\textbf{40} & 41/\textbf{40} & 40/\textbf{40} & 40/\textbf{40} & 40/\textbf{40} & 40/\textbf{40} & 40/\textbf{40} & 59/\textbf{60} & 61/\textbf{60} & 59/\textbf{59} & 61/\textbf{61} & 559/\textbf{559} \\
\bottomrule
\end{tabular}
}
\end{table}

\section{Results Details}

\subsection{Example Compiled Puzzle Prompts (Gemini 3.5 Flash Lite)}
\label{sec:example-prompts}

We show the instruction and two of four complete demonstrations from the seed-42 \textsc{MIPROv2} prompt, followed by the complete static portion of the seed-42 \textsc{SIMBA} prompt. \promptformat{Green} text denotes output-format constraints, \promptchess{orange} text denotes chess guidance, \promptexample{blue} text denotes optimizer-selected demonstrations, and \promptother{pink} text denotes other content. The full seed-42 \textsc{MIPROv2} prompt contains 1,507 characters; the 1,449-character value in Table~\ref{tab:puzzle-prompt-size} is the rounded mean over seeds 42--44 (1,507, 1,514, and 1,327 characters). The complete \textsc{SIMBA} prompt below contains 901 characters; its corresponding 930-character table value averages 901, 1,009, and 879 characters.

\begin{promptbox}[\textsc{MIPROv2} seed 42: instruction + two of four demonstrations]
\small\ttfamily
\promptchess{Analyze the given board position (\textsc{FEN}) and determine the absolute best tactical move for the active side.}\par
\promptformat{Select exactly one move from the provided list of legal moves (legal\_moves\_uci). Your response must only contain the selected move with no reasoning, explanations, or labels.}\par\smallskip
\promptexample{Position Fen: 8/5pkp/2BR2p1/1pP1p3/1P5q/3QP2P/5P1K/r7 b - - 0 32\newline
Side To Move: black\newline
Legal Moves Uci: g7h8, g7g8, g7f8, g7h6, h4d8, h4e7, h4h6, h4f6, h4h5, h4g5, h4g4, h4f4, h4e4, h4d4, h4c4, h4b4, h4h3, h4g3, h4f2, a1a8, a1a7, a1a6, a1a5, a1a4, a1a3, a1a2, a1h1, a1g1, a1f1, a1e1, a1d1, a1c1, a1b1, h7h6, f7f6, g6g5, e5e4, h7h5, f7f5\newline
Move: h4f2}\par\smallskip
\promptexample{Position Fen: 4r1k1/R5pp/2p5/8/2P2p2/1PNn1P2/P2P2PP/6K1 b - - 4 31\newline
Side To Move: black\newline
Legal Moves Uci: g8h8, g8f8, e8f8, e8d8, e8c8, e8b8, e8a8, e8e7, e8e6, e8e5, e8e4, e8e3, e8e2, e8e1, d3e5, d3c5, d3b4, d3f2, d3b2, d3e1, d3c1, h7h6, g7g6, c6c5, h7h5, g7g5\newline
Move: e8e1}\par
\promptexample{[Two additional optimizer-selected demonstrations.]}
\end{promptbox}

\begin{promptbox}[\textsc{SIMBA} seed 42: complete 901-character static prompt]
\small\ttfamily
\promptchess{Solve the chess puzzle by finding the single best move.}\par
\smallskip
\promptexample{Position Fen: r1b2Rnk/2p3pB/2p4p/1pq1P1N1/p7/R1P1P3/5PPP/6K1 w - - 1 29\newline
Side To Move: white\newline
Legal Moves Uci: f8g8, f8e8, f8d8, f8c8, f8f7, f8f6, f8f5, f8f4, f8f3, h7g8, h7g6, h7f5, h7e4, h7d3, h7c2, h7b1, g5f7, g5e6, g5e4, g5h3, g5f3, a3a4, a3b3, a3a2, a3a1, g1h1, g1f1, e5e6, e3e4, c3c4, h2h3, g2g3, f2f3, h2h4, g2g4, f2f4\newline
Move: f8g8}\par\smallskip
\promptexample{Position Fen: 8/5pk1/p3pNp1/1p2P3/6P1/2Q2Pqr/2R1K3/8 b - - 8 37\newline
Side To Move: black\newline
Legal Moves Uci: g7h8, g7f8, g7h6, h3h8, h3h7, h3h6, h3h5, h3h4, h3h2, h3h1, g3e5, g3h4, g3g4, g3f4, g3f3, g3h2, g3g2, g3f2, g3g1, g3e1, g6g5, a6a5, b5b4\newline
Move: g3g2}\par\smallskip
\promptexample{Position Fen: 3k4/p4Rp1/p7/6p1/Q2PP3/6r1/P5q1/3RK3 b - - 5 26\newline
Side To Move: black\newline
Legal Moves Uci: d8c8, g3g4, g3h3, g3f3, g3e3, g3d3, g3c3, g3b3, g3a3, g2e4, g2h3, g2f3, g2h2, g2f2, g2e2, g2d2, g2c2, g2b2, g2a2, g2h1, g2g1, g2f1, g7g6, a6a5, g5g4\newline
Move: Re3+}
\end{promptbox}

\subsection{Puzzle Move Quality by Regret}
\label{sub:regret}

While puzzle accuracy gives insight on performance, the outcome is binary success / failure. A decent human player can play good moves without producing the best one in puzzles.
To give a more lenient perspective on move quality, we evaluate the regret in the puzzle moves for baseline and best prompt in long puzzles (3--5 moves) in Table~\ref{tab:position-regret}. GPT-5.6 Luna is omitted because its selected prompt is unchanged from the baseline.
The regret is estimated as the difference between the \textsc{Stockfish} evaluation of the best move and the move played by the model.
With regret the move is not binary evaluated but offers a more granular perspective on how bad an incorrect move is.

\begin{table}[h]
\centering
\caption{Regret computed on long puzzles (3--5 moves) with \textsc{Stockfish} depth 20. Rank entries report median-regret rank / position-accuracy rank / puzzle-accuracy rank; lower ranks are better. Ranks are among the five models shown, separately for the Baseline and Best columns; position accuracy is measured on the same 3--5-move positions, and puzzle accuracy is taken from Table~\ref{tab:puzzle-optim-results}. Bottom row is the meta-model.}
\label{tab:position-regret}
\scriptsize
\setlength{\tabcolsep}{3pt}
\resizebox{0.7\textwidth}{!}{\begin{tabular}{llrrrrrr}
\toprule
\multicolumn{2}{c}{Model} & \multicolumn{2}{c}{Mean regret (cp)} & \multicolumn{2}{c}{Median regret (cp)} & \multicolumn{2}{c}{Rank (regret / position / puzzle)} \\
\cmidrule(lr){1-2}\cmidrule(lr){3-4}\cmidrule(lr){5-6}\cmidrule(lr){7-8}
& Best method & Baseline & Best & Baseline & Best & Baseline & Best \\
\midrule
Gemini 3.5 Flash Lite & \textsc{SIMBA} & 49,964 & 39,763 & 1,059 & 568 & 2 / 2 / 2 & 1 / 1 / 1 \\
GPT-4o Mini & \textsc{SIMBA} & 71,662 & 70,134 & 100,265 & 100,202 & 5 / 5 / 5 & 5 / 5 / 5 \\
Qwen3.8 27B & \textsc{GEPA} & 70,249 & 69,169 & 100,179 & 100,179 & 4 / 4 / 4 & 4 / 4 / 4 \\
DeepSeek V4 Pro 0813 & \textsc{GEPA} & 66,216 & 61,755 & 5,723 & 1,935 & 3 / 3 / 3 & 3 / 3 / 3 \\
Muse Spark 1.2 Contributor & \textsc{SIMBA} & 49,376 & 40,268 & 830 & 834 & 1 / 1 / 1 & 2 / 2 / 2 \\
\midrule
Gemini 3.5 Flash & --- & 19,117 & --- & 30 & --- & --- & --- \\
\bottomrule
\end{tabular}
}
\end{table}
The results are gathered in Table~\ref{tab:position-regret}, together with each model's regret, position-accuracy, and puzzle-accuracy ranks. 
Position accuracy treats every move of every puzzle independently, whereas a five-move puzzle requires all five positions to be correctly predicted to count as correct.
Note that regret can be very harsh if the model misses a checkmate or gets checkmated; for example, missing a checkmate or predicting an illegal move can be penalized by 200,000 centipawns, while for reference the value of a queen is estimated at 900 centipawns. Hence, we report median regret to reduce the impact of such penalties.
Median regret, position accuracy, and puzzle accuracy rank the five models identically within each column of Table~\ref{tab:position-regret}, although the gaps between models differ across metrics. This agreement supports using regret as a complementary measure of move quality; it does not validate it as a general measure of playing strength. In game-play rollouts (Section~\ref{sub:playing-chess-transfer}), regret remains applicable without a unique reference move.

\subsection{Prompt Size}
\label{sec:prompts-size}

Table~\ref{tab:puzzle-prompt-size} reports the average compiled-prompt size in characters. The base prompt is 55 characters. Some algorithms never manage to improve the base prompt in some models, showing that the task is not straightforward to improve. Across the evaluated model--optimizer seeds in Table~\ref{tab:puzzle-optim-results}, 21 of 141 runs (14.9\%) leave the prompt unchanged.

\begin{table*}[!ht]
\centering
\caption{\textbf{Average prompt size in characters, over three seeds.} The base prompt is 55 characters.
Some algorithms never manage to improve the base prompt on some models (gray cells).
Bold marks the optimizer with the highest puzzle accuracy for each model. N.E. marks not evaluated.}
\label{tab:puzzle-prompt-size}
\scriptsize
\setlength{\tabcolsep}{5pt}
\begin{tabular}{lrrrrrr}
\toprule
Model & \textsc{BFS} & \textsc{BRS} & \textsc{COPRO} & \textsc{GEPA} & \textsc{MIPROv2} & \textsc{SIMBA} \\
\midrule
Gemini 3.5 Flash Lite & 4,497 & 4,873 & 336 & 1,725 & 1,449 & \textbf{930} \\
GPT-4o Mini & 4,407 & 4,671 & 367 & 3,225 & 1,423 & \textbf{579} \\
Jev 1.13 & 4,506 & 4,789 & 921 & \textbf{2,968} & 1,504 & N.E. \\
Qwen3.8 27B & 4,588 & 4,604 & 227 & \textbf{2,109} & 1,446 & 1,050 \\
Claude Haiku 4.5 & 4,565 & 4,506 & \cellcolor{gray!20}55 & \textbf{3,626} & 1,070 & 1,157 \\
GPT-5.6 Luna & 4,593 & \cellcolor{gray!20}\textbf{55} & 146 & 3,045 & 154 & \cellcolor{gray!20}55 \\
DeepSeek V4 Pro 0813 & 4,662 & 4,885 & 239 & \textbf{3,057} & 1,332 & 1,038 \\
Muse Spark 1.2 Contributor & 4,577 & 4,875 & \cellcolor{gray!20}55 & 768 & 1,542 & \textbf{929} \\
\bottomrule
\end{tabular}
\end{table*}

\subsection{Task Prompts and Signatures}
\label{sec:task-prompts}

All generative-model results use the \textbf{JSON adapter} (\textsc{DSPy}'s \texttt{JSONAdapter}), which enforces strict structured output formatting. Jev 1.13 is the exception: it is queried through a decision endpoint with its own schema adapter, not through \textsc{DSPy}'s generative JSON adapter.

\paragraph{Puzzles (Independent).} The default instruction is: \emph{``Solve the chess puzzle by finding the single best move.''} This 55-character instruction is distinct from the adapter-generated formatting instructions and input fields.
The input fields are:
\begin{itemize}[nosep]
    \item \texttt{position\_fen}: current position in \textsc{FEN}.
    \item \texttt{side\_to\_move}: \texttt{white} or \texttt{black}.
    \item \texttt{legal\_moves\_uci}: comma-separated legal \textsc{UCI} moves.
\end{itemize}
The output field is:
\begin{itemize}[nosep]
    \item \texttt{move}: exactly one legal \textsc{SAN} or \textsc{UCI} move, without reasoning or labels.
\end{itemize}

The resulting generative-model request contains the following fixed \textsc{JSONAdapter} scaffolding around the optimized instruction and puzzle-specific fields. 
\textsc{JSONAdapter} is proposed by DSPY \cite{khattab2024dspy} to structure inputs.
Placeholders in angle brackets are replaced at runtime.

\begin{promptbox}[Baseline JSONAdapter request template]
\begin{promptverb}
SYSTEM:
Your input fields are:
1. `position_fen` (str): Current position in FEN
2. `side_to_move` (str): white or black
3. `legal_moves_uci` (str): Comma-separated legal UCI moves
Your output fields are:
1. `move` (str): Exactly one legal SAN or UCI move, without reasoning or labels
All interactions will be structured in the following way, with the appropriate values filled in.

Inputs will have the following structure:

[[ ## position_fen ## ]]
{position_fen}

[[ ## side_to_move ## ]]
{side_to_move}

[[ ## legal_moves_uci ## ]]
{legal_moves_uci}

Outputs will be a JSON object with the following fields.

{
  "move": "{move}"
}
In adhering to this structure, your objective is:
        <<PROMPT>>

USER:
[[ ## position_fen ## ]]
<<FEN>>

[[ ## side_to_move ## ]]
<<SIDE_TO_MOVE>>

[[ ## legal_moves_uci ## ]]
<<LEGAL_MOVES_UCI>>

Respond with a JSON object in the following order of fields: `move`.
\end{promptverb}
\end{promptbox}

\subsection{Interpreting \textsc{Lichess} Puzzle Ratings for LLMs}
\label{sub:puzzle-difficulty}

Our benchmark uses the \textsc{Lichess} puzzle rating as a difficulty axis. 
This rating is a human-calibrated scale, so we first check what it means for LLMs.

The puzzle rating is derived from \textsc{Lichess}'s puzzle-difficulty classification, which uses the \textsc{Glicko-2} system~\cite{glickman2001dynamic, glickman2012example}. \textsc{Glicko-2} is a paired-comparison rating system: like Elo, it models the expected outcome of a comparison as a function of the relative ratings of the two entities. In the \textsc{Lichess} model, a puzzle is treated analogously to an opponent that wins whenever the user fails to solve it.

The key difference between the two systems is that \textsc{Glicko-2} also accounts for uncertainty in the ratings.

By definition, these ratings are relative and impose a scalar ordering based on the performance of users on the platform, whether the rated entities are human players or puzzles.
However, constructing a benchmark of chess puzzles requires us to freeze the puzzle ratings and ignore the uncertainty aspect of \textsc{Glicko-2} and interpret ratings as an external descriptive variable. 

While the puzzle rating is calibrated for human players, no such calibration exists for LLMs. Consequently, \textbf{we ask whether it can serve as a proxy for task difficulty for LLMs in our benchmark}.

Empirically, we observe significant correlations between LLMs' puzzle solve rates and puzzle ratings. Across pooled 200-point Elo bins, correlations were \(r=-0.937\) (\(p<0.001\), 8 bins) and \(r=-0.996\) (\(p=0.004\), 4 bins) for 1--2- and 3--5-move puzzles, respectively.
Each 200-point Elo increase multiplies solve odds by 0.72 for 1–2-move puzzles and 0.45 for 3–5-move puzzles, corresponding to odds reductions of 28\% and 55\%, respectively. Under the standard Elo model, a 200-point disadvantage multiplies the odds by \(10^{-200/400}\approx0.316\), corresponding to a reduction of 68.4\% odds. 
For an equal-rated player, this changes the expected success rate from 50\% to approximately 24\%, a decrease of 26 percentage points, or 52\% relative to the original success rate. 
Notice that this is not the rating-update rule used by \textsc{Lichess}; it provides a simple theoretical reference for interpreting rating differences.

\begin{figure*}[!ht]
    \centering
    \includegraphics[width=0.7\textwidth]{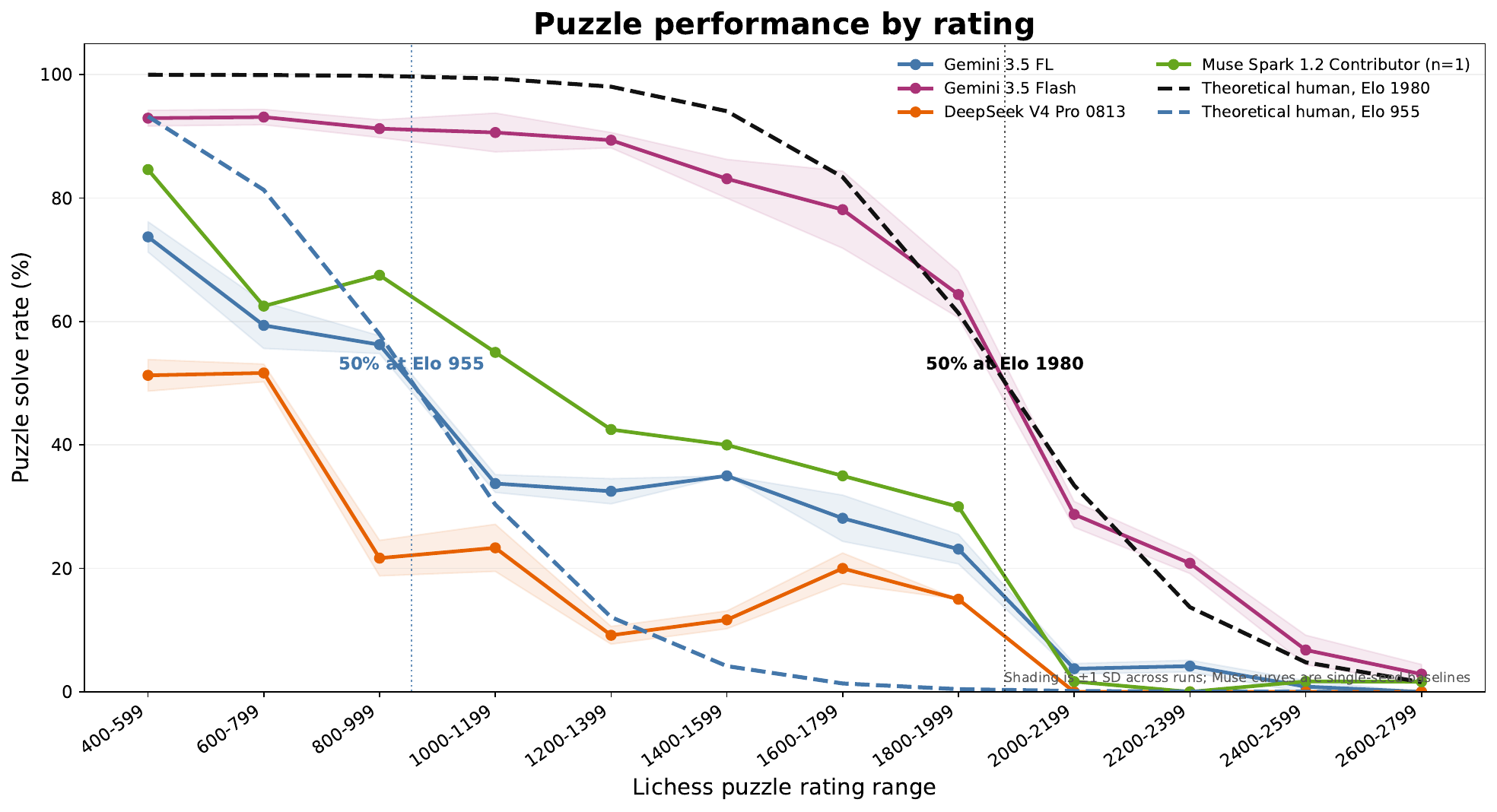}
    \caption{\textbf{Puzzle solve rate by \textsc{Lichess} rating with the baseline prompt.}
    Solid curves show empirical model performance; shaded regions show $\pm 1$ standard deviation across runs. 
    Dashed curves show the standard Elo success-rate function for hypothetical players anchored at $50\%$ success for puzzle ratings 955 and 1980. 
    The vertical lines mark these anchor ratings.}
    \label{fig:puzzle-elo-focused}
\end{figure*}

Figure~\ref{fig:puzzle-elo-focused} reports puzzle solve rates by rating for the models evaluated. 
The solid curves show the performance of the empirical model. For reference, the dashed curves show the success rate predicted by the standard Elo model for hypothetical players whose ratings are chosen so that their expected success rate is $50\%$ at puzzle ratings 955 and 1980, respectively. 
These anchor points approximately correspond to the observed $50\%$ solve-rate points of Gemini 3.5 Flash Lite and Gemini 3.5 Flash. 
The vertical lines mark the two corresponding anchor ratings.

\subsection{Hyperparameters}
\label{sec:hps}

\paragraph{LM configuration.}

\begin{table}[h]
\centering
\caption{\textbf{Exact OpenRouter model identifiers used in the puzzle experiments.} Gemini 3.5 Flash is also the fixed reflection and meta-model.}
\label{tab:openrouter-model-identifiers}
\scriptsize
\begin{tabular}{ll}
\toprule
Model & OpenRouter identifier \\
\midrule
Gemini 3.5 Flash & \texttt{openrouter/google/gemini-3.5-flash} \\
Gemini 3.5 Flash Lite & \texttt{openrouter/google/gemini-3.5-flash-lite} \\
GPT-4o Mini & \texttt{openrouter/openai/gpt-4o-mini} \\
Jev 1.13 & \texttt{typesafe/jev-1.13-20260917} \\
Qwen3.8 27B & \texttt{openrouter/qwen/qwen3.8-27b} \\
Claude Haiku 4.5 & \texttt{openrouter/anthropic/claude-haiku-4.5} \\
GPT-5.6 Luna & \texttt{openrouter/openai/gpt-5.6-luna} \\
DeepSeek V4 Pro 0813 & \texttt{openrouter/deepseek/deepseek-v4-pro-0813} \\
Muse Spark 1.2 Contributor & \texttt{openrouter/meta/muse-spark-1.2-contributor} \\
\bottomrule
\end{tabular}
\end{table}

\begin{table}[h]
\centering
\caption{\textbf{Language model configuration for compilation and evaluation.}}
\label{tab:lm-config}
\begin{tabular}{llccc}
\toprule
Role & Model & Temp & Max tokens & Provider \\
\midrule
Baseline task LM & (per-model) & 0 & Task-specific & Mixed \\
Optimization task LM & (per-model, Sec~\ref{sec:model-selection}) & 0 & (per-model)  & OpenRouter \\
Meta-model & \texttt{gemini-3.5-flash} & 1 & 8192 & OpenRouter \\
Prompt LM (\textsc{COPRO}, $n>1$) & \texttt{gemini-3.5-flash} & 1 & 8192 & OpenRouter \\
\bottomrule
\end{tabular}
\end{table}

\begin{sloppypar}
For the optimization panel, task LMs use \texttt{temperature=0.0}, while the meta-model and \textsc{COPRO} prompt LM use \texttt{temperature=1.0}.
All selected LMs use \texttt{max\_concurrency=1}, \texttt{retry\_backoff\_seconds=0.25}, \texttt{cache\_mode=stateless}, \texttt{drop\_params=true}. The timeout is 30\,s, except for Qwen3.8 27B and DeepSeek V4 Pro 0813 (120\,s), GPT-5.6 Luna (300\,s), and Muse Spark 1.2 Contributor (600\,s). Runs use \texttt{max\_retries=1}.
Gemini 3.5 Flash Lite additionally uses \texttt{capture\_reasoning=true}, \texttt{reasoning\_effort=low}, and \texttt{include\_reasoning=true}; 
Gemini 3.5 Flash used default reasoning with \texttt{capture\_reasoning=false}.
GPT-4o Mini uses \texttt{capture\_reasoning=false} and \texttt{reasoning\_effort=none}; Claude Haiku 4.5, Qwen3.8 27B, and DeepSeek V4 Pro 0813 use the provider's default reasoning.
Jev 1.13 is evaluated through the Typesafe decision endpoint (\texttt{/api/alpha/decisions}) rather than a generative OpenRouter chat endpoint. The decision request carries the same information as the generative prompts: the instructions and demonstrations, the position fields (\textsc{FEN} and side to move), and the legal moves, which serve as the answer options. The model does not generate text. It chooses among the listed legal moves only, so its answer space is exactly the provided option set. The endpoint bills input tokens only and exposes no temperature or sampling controls. Generative settings such as \texttt{max\_output\_tokens}, \texttt{temperature}, and the JSON adapter therefore do not apply. The absence of sampling control is also why \textsc{SIMBA}, which queries the model at several temperatures, is not evaluated for Jev 1.13 (N.A. in Table~\ref{tab:puzzle-optim-results}).
\end{sloppypar}

\textbf{Engine configuration.} Puzzle regret and game-play rollouts (Appendix~\ref{sub:regret} \& \ref{sub:playing-chess-transfer}) use \textsc{Stockfish} 18 at depth 20.

\paragraph{Optimizer hyperparameters.} Table~\ref{tab:optimizer-hyperparameters} lists the hyperparameters for all optimizers.

\begin{table}[!ht]
\centering
\caption{\textbf{Optimizer hyperparameters.} Unless noted otherwise, values are \textsc{DSPy} defaults.}
\label{tab:optimizer-hyperparameters}
\small
\begin{tabular}{lll}
\toprule
Optimizer & Parameter & Value \\
\midrule
\textsc{GEPA} & \texttt{auto} & \texttt{heavy} ($\sim$3,436 metric calls) \\
 & \texttt{candidate\_selection\_strategy} & \texttt{pareto} \\
 & \texttt{use\_merge} & \texttt{True} \\
 & \texttt{skip\_perfect\_score} & \texttt{True} \\
 & \texttt{reflection\_minibatch\_size} & 3 \\
 & \texttt{num\_threads} & 16 \\
\midrule
\textsc{MIPROv2} & \texttt{auto} & \texttt{heavy} \\
 & \texttt{minibatch} & \texttt{False} \\
 & \texttt{max\_bootstrapped\_demos} & 4 \\
 & \texttt{max\_labeled\_demos} & 4 \\
 & \texttt{num\_threads} & 16 \\
 & \texttt{num\_candidates} & $\sim$18 (set by \texttt{auto=heavy}) \\
 & \texttt{num\_trials} & $\sim$27 (set by \texttt{auto=heavy}) \\
 & \texttt{program\_aware\_proposer} & \texttt{True} \\
 & \texttt{data\_aware\_proposer} & \texttt{True} \\
 & \texttt{tip\_aware\_proposer} & \texttt{True} \\
 & \texttt{fewshot\_aware\_proposer} & \texttt{True} \\
 & \texttt{prompt\_max\_tokens} & 8192 \\
\midrule
\textsc{SIMBA} & \texttt{bsize} & 32 \\
 & \texttt{num\_candidates} & 6 \\
 & \texttt{max\_steps} & 8 \\
 & \texttt{max\_demos} & 4 \\
 & \texttt{num\_threads} & 16 \\
\midrule
\textsc{COPRO} & \texttt{breadth} & 10 \\
 & \texttt{depth} & 3 \\
 & \texttt{eval\_num\_threads} & 16 \\
\midrule
\textsc{BFS} & \texttt{max\_bootstrapped\_demos} & 4 \\
 & \texttt{max\_labeled\_demos} & 16 \\
 & \texttt{max\_rounds} & 1 \\
\midrule
\textsc{BRS} & \texttt{max\_bootstrapped\_demos} & 4 \\
 & \texttt{max\_labeled\_demos} & 16 \\
 & \texttt{num\_candidate\_programs} & 16 \\
 & \texttt{num\_threads} & 16 \\
 & \texttt{max\_rounds} & 1 \\
\bottomrule
\end{tabular}
\end{table}

\end{document}